\documentclass[runningheads]{llncs}

\usepackage[mobile]{eccv}

\usepackage{eccvabbrv}

\usepackage{graphicx}
\usepackage{booktabs}
\usepackage{ulem}

\usepackage[accsupp]{axessibility}  % Improves PDF readability for those with disabilities.

\usepackage{hyperref}

\usepackage{orcidlink}
\usepackage{makecell}
\usepackage{color}
\usepackage{multirow}
\usepackage{float}
\newcommand\networkname{PBaSR}

\newcommand\best[1]{\textcolor{black}{\textbf{#1}}}
\newcommand\second[1]{\textcolor{red}{\textbf{#1}}}

\newfloat{figtab}{htb}{fgtb}
\makeatletter
  \newcommand\figcaption{\def\@captype{figure}\caption}
  \newcommand\tabcaption{\def\@captype{table}\caption}
\makeatother

\begin{document}

\newcommand\oldtitle{Exploring Blind Super-Resolution for Real-world Blur Images}
\newcommand\newtitle{A New Dataset and Framework for Real-World Blurred Images Super-Resolution}

\title{\newtitle}

\titlerunning{Real-World Blurred Images Super-Resolution}

\renewcommand{\thefootnote}{\fnsymbol{footnote}}
\author{Rui Qin\inst{1,2}\orcidlink{0000-0001-5901-9775} \and
Ming Sun\inst{2}\orcidlink{0000-0002-5948-2708} \and
Chao Zhou\inst{2} \and
Bin Wang\inst{1}$^{,\dag}$\orcidlink{0000-0002-5176-9202}
}
\footnotetext[4]{Corresponding author.}

\authorrunning{R. Qin et al.}

\institute{School of Software, Tsinghua University, Beijing, China \\
\email{qr20@mails.tsinghua.edu.cn}, \email{wangbins@tsinghua.edu.cn} \and
Kuaishou Technology, Beijing, China \\
\email{\{sunming03, zhouchao\}@kuaishou.com}}

\maketitle

\begin{abstract}
    Recent Blind Image Super-Resolution (BSR) methods have shown proficiency in general images. However, we find that the efficacy of recent methods obviously diminishes when employed on image data with blur, while image data with intentional blur constitute a substantial proportion of general data. To further investigate and address this issue, we developed a new super-resolution dataset specifically tailored for blur images, named the Real-world Blur-kept Super-Resolution (ReBlurSR) dataset, which consists of nearly 3000 defocus and motion blur image samples with diverse blur sizes and varying blur intensities. Furthermore, we propose a new BSR framework for blur images called Perceptual-Blur-adaptive Super-Resolution (PBaSR), which comprises two main modules: the Cross Disentanglement Module (CDM) and the Cross Fusion Module (CFM). The CDM utilizes a dual-branch parallelism to isolate conflicting blur and general data during optimization. The CFM fuses the well-optimized prior from these distinct domains cost-effectively and efficiently based on model interpolation. By integrating these two modules, PBaSR achieves commendable performance on both general and blur data without any additional inference and deployment cost and is generalizable across multiple model architectures. Rich experiments show that PBaSR achieves state-of-the-art performance across various metrics without incurring extra inference costs. Within the widely adopted LPIPS metrics, PBaSR achieves an improvement range of approximately 0.02-0.10 with diverse anchor methods and blur types, across both the ReBlurSR and multiple common general BSR benchmarks. \href{https://github.com/Imalne/PBaSR}{Code here.}
\end{abstract}
% !TEX root = main.tex

\section{Introduction}
\label{sec:intro}

\begin{figure}[t]
    \centering
    \includegraphics[width=0.9\linewidth]{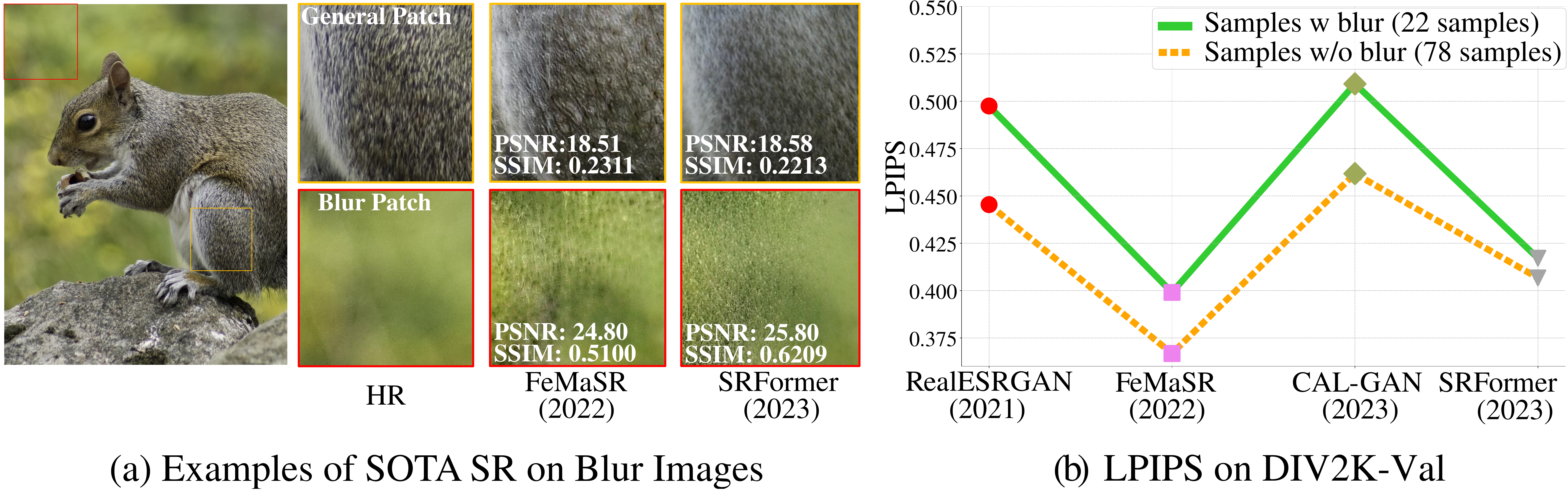}
    \caption{Recent BSR methods' performance on blur data. \textbf{Left}: Examples of real images with blur. \textbf{Right}: LPIPS on blur and non-blur images in DIV2K-Val.
    }
    \label{fig:fig1}
\end{figure}

Blind Image Super-Resolution (BSR)~\cite{realesrgan_wang2021real,bsrgan_zhang2021designing,liang2021swinir,mou2022mmrealsr,femasr_chen2022real,zhou2023srformer,chen2023activating,qin2023blind,chen2024cassr,chen2024ntire} aims to realistically reconstruct High-Resolution (HR) images from Low-Resolution (LR) images with unknown degradation. Benefiting from the construction of various benchmarks~\cite{lim2017enhanced_flickr2k,Agustsson_2017_div2k,gu2019div8k,Set14,Set5,manga109,wu2022animesr}, the BSR technique has achieved considerable success in the field of general images~\cite{gu2019blind,ji2020real,liang2021swinir,huang2020unfolding,pan2021exploiting,wang2021unsupervised} 
as well as in various specific image categories, such as face images~\cite{chen2021progressive,li2020blind,li2020enhanced}, cartoon images~\cite{wu2022animesr}, text images~\cite{qin2022scene,Chen_2021_CVPR,PCAN2021}, and remote sensing images~\cite{xiao2023degrade,gong2021enlighten}.

Blurring, a prevalent form of image degradation, is influenced by factors such as aperture size and exposure time, manifesting in actual images with different visual perceptions. Previous research on blur often treated it negatively, focusing on its detection~\cite{tang2020defusionnet,jin2023depth} and removal~\cite{cho2021rethinking,abuolaim2020defocus}. 
However, upon investigation~\cite{adobe_Survey2014,goolge_survey}, we found that blurring is also a commonly used visual perception enhancement photography technique. Take the real sample from DIV2K~\cite{Agustsson_2017_div2k} in \cref{fig:fig1}a as an example, it accentuates the foreground squirrel by defocusing the background. In addition, this type of technique is broadly utilized in numerous commercial multimedia editing platforms, evidenced by the `Blur' features within the basic `Filter' menus of Adobe Photoshop, Krita, and LunaPic.
Therefore, 
    we believe that the intentionally set blur should be preserved rather than entirely removed during the BSR process, which has not been paid much attention to yet. As shown in~\cref{fig:fig1}a, recent methods over-enhance the blur region despite the overall high PSNR/SSIM, which impairs the actual perception quality. 
To quantitatively verify our surmise, 
    we also used a more reasonable perceptual metric, LPIPS~\cite{LPIPS_zhang2018unreasonable}, for a brief evaluation on the DIV2K-Val~\cite{Agustsson_2017_div2k} benchmark (100 samples)  and split it into samples with blur vs. those without blur. As shown in \cref{fig:fig1}b, the LPIPS of blur samples generally decreases by 0.01$\sim$0.06 across different methods compared to the fully-focused samples without blur.
To further investigate this neglected problem, 
    we develop a new super-resolution benchmark specifically tailored for the restoration of images with blur, named the Real-world Blur-kept Super-Resolution (ReBlurSR) dataset. It contains 2,931 high-quality (HQ) images with diverse blur sourced from the existing super-resolution (\textit{e.g.}, DIV2K, DIV8K~\cite{gu2019div8k}, Flickr2K~\cite{lim2017enhanced_flickr2k}), blur-related datasets (\textit{e.g.}, CUHK~\cite{CUHK}, EBD~\cite{EBD}) and web images. We meticulously labeled and categorized the blur data among various characteristics, such as types, sizes, and intensities. 

Except for the ReBlurSR dataset, we argue that an optimal blur BSR framework should adaptively handle blur images and enhance their restoration performance without affecting general image processing quality. Furthermore, it should integrate easily with existing SOTA models, with minimal additional inference costs. In response to these criteria, we propose the Perceptual-Blur-adaptive Super-Resolution (PBaSR), the first blur-adaptive BSR framework.
The PBaSR framework consists of two main modules: the Cross Disentanglement Module (CDM) and the Cross Fusion Module (CFM). On the one hand, the CDM, with its branch tailored for blur, separates blur from non-blur data for focused learning and reduces conflicts in uniform training. On the other hand, the CFM utilizes adaptive cross-branch weight interpolation, fostering the exchange of information between branches while keeping feature spaces aligned. This setup allows for the effective fusion of optimal priors from each branch in its respective domains through cost-efficient weight averaging, thereby handling both blur and non-blur data well without adding any inference or deployment costs. By synergizing these modules, PBaSR substantially elevates the performance of various recent mainstream BSR methods on blur data.
Comprehensive testing on ReBlurSR and several general BSR benchmarks with various quantitative metrics shows that PBaSR achieves state-of-the-art performance in blur image blind super-resolution and matches recent top-performing methods for general data processing. PBaSR consistently enhances performance across different perceptual metrics, suitable for various methods and blur types, showing a 0.02 to 0.10 improvement in LPIPS without any additional inference costs.
In summary, our main contributions are as follows:
\begin{itemize}
    \item We observed that recent BSR methods ignore the treatment of images with intentional blur, and have limitations on the preservation of blur regions.
    \item In response to this phenomenon, we propose the Real-world Blur-kept Super-Resolution (ReBlurSR) dataset for blur image blind super-resolution. It contains 2,931 high-quality images with diverse blur regions, nearly three times the size of the commonly used SR benchmark DIV2K. 
    \item We validate the limitations of current methods on blur BSR, and propose the new Perceptual-Blur-adaptive Super-Resolution (PBaSR) framework for blur BSR.
    Extensive experiments on multiple benchmarks show that PBaSR effectively improves BSR performance on blur data while maintaining commendable performance on general data, with no additional inference cost.
\end{itemize}
% !TEX root = main.tex

\section{Related Work}

\subsection{Single Image Super-Resolution}
Since the advent of SRCNN~\cite{srcnndong2015image}, many CNN-based super-resolution frameworks~\cite{kim2016deeply,dong2014learning,shi2016real} have been developed. EDSR~\cite{lim2017enhanced_flickr2k} and RDN~\cite{rdnzhang2018residual} improved SISR using Residual Dense Blocks~\cite{huang2017densely}. Recently, frameworks using VQVAE~\cite{femasr_chen2022real,qin2023blind,zhou2022towards} and Transformer~\cite{zhou2023srformer,liang2021swinir,liu2023reconstructed,qu2024xpsr} have shown better results. BSR is particularly challenging due to diverse degradation, leading many to use GANs~\cite{wang2018esrgan,menon2020pulse,zhu2017unpaired} for enhancing texture and incorporating well-designed degradation synthesis processes~\cite{bsrgan_zhang2021designing,realesrgan_wang2021real} with high-quality datasets. As noted in \cref{sec:intro}, intentional blur can improve visual perception~\cite{adobe_Survey2014,goolge_survey}. \cref{fig:fig1} shows that over 20\% of DIV2K-Val samples are blurred, and current methods struggle with such data. Thus, researching blur BSR and constructing datasets is beneficial for BSR techniques.

\subsection{Image Super-Resolution Benchmark Datasets}
Recent image super-resolution (SR) has been significantly driven by the development of comprehensive benchmarks. The DIV2K~\cite{Agustsson_2017_div2k}, including 1000 high-quality (HQ) 2K resolution images, has become a widely used standard for evaluating SR models due to its diverse and detailed content. Similarly, the Flickr2K~\cite{lim2017enhanced_flickr2k}, with its 2650 HQ images from various scenarios, provides sufficient data for training robust SR models. For evaluation, the BSDS100~\cite{BSDS100} is widely used, with 100 challenging images with complex natural scenes. The Set5~\cite{Set5} and Set14~\cite{Set14} datasets, despite their smaller sizes, remain popular for their historical significance. Besides, the Urban100~\cite{urban100} and Manga109~\cite{manga109} datasets cater to specific needs, with the former focusing on urban scenes and the latter on comic images. Though these datasets provide robust platforms for the evaluation of general SR tasks, there is still a lack of specialized benchmarks for blur image SR.

% !TEX root = main.tex

\section{Methodology}
\label{sec:method}

\subsection{Blur in Real-World Blind Super-Resolution}
Though recent BSR methods excel in general image enhancement, they do not exhibit optimal performance when facing blur images, while blur is widespread in high-quality multimedia content and is an important photographic technique.

To illustrate this, we selected blur samples from the DIV2K-val dataset, revealing recent SR methods' limitations, such as over-texturization and vignetting destruction, as shown in \cref{fig:fig1}a. These problems are overlooked by PSNR/SSIM metrics due to their insensitivity to smooth areas. To further assess the inadequacy quantitatively, we conducted statistical experiments using the DIV2K-Val dataset. Specifically, we used the blur detection method D-DFFNet~\cite{jin2023depth} to differentiate samples into those with and without blur. We found that over 20\% of samples had noticeable blur, which is a non-negligible proportion. Besides, as \cref{fig:fig1}b indicates, the SOTA BSR methods showed a marked decrease in effectiveness on the blur samples, with a 0.01-0.06 drop in LPIPS.

These findings highlight blur's significance in high-quality media, a gap in current datasets and BSR methods. Therefore, this work aims to develop a focused subset for this gap and improve BSR methods' performance on blur data.

\subsection{Real-World Blur-Kept Super-Resolution Dataset}
To support the training and evaluation of BSR methods on blur images, we created the high-quality \textbf{Re}al-world \textbf{Blur}-kept \textbf{S}uper-\textbf{R}esolution dataset, namely, ReBlurSR dataset, comprising of two subsets. We selected 2210 real-world blur images from sources like DIV2K, Flickr2K, DIV8K, CUHK, EBD, and the web. Additionally, we synthesized 601 blur images using popular diffusion-based models. These 2811 images form the ReBlurSR-Train set. For realistic evaluation, the ReBlurSR-Test set includes 120 real blur images from existing validation benchmarks. Each sample has a High-Resolution (HR) image and a blur region map, indicating blur pixels with 0 and non-blur pixels with 1.
\begin{figure*}[t]
    \centering
    \includegraphics[width=\linewidth]{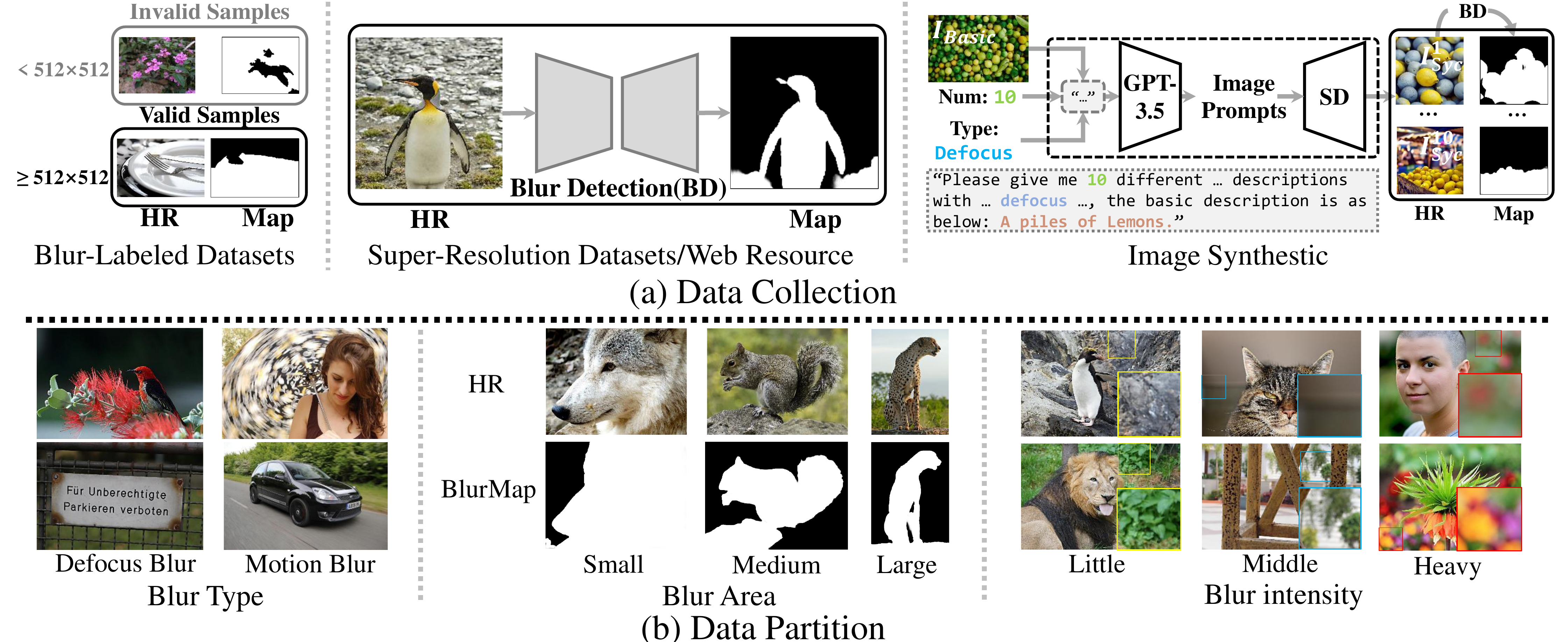}
    \caption{Data collection and partition of ReBlurSR. (a) Three methods of collecting the ReBlurSR dataset. (b) The samples from different data partitions. 
    }
    \label{fig:data_split}
\end{figure*}

\noindent\textbf{Data Collection}
For diversity and realism, we gathered data from recent public datasets, the web and synthesized realistic blur images using multi-modal models and image-generating models. As shown in \cref{fig:data_split}a, our data collection involved:

\textit{1)} From blur-specific datasets EBD~\cite{EBD} and CUHK~\cite{CUHK}, we chose large samples (over 512$\times$512 pixels) with less than 80\% blur region and paired the provided manually labeled blur maps with images, resulting in 849 samples.

\textit{2)} Using D-DFFNet~\cite{jin2023depth} and method in~\cite{kim2018defocus} for blur estimation on universal SR datasets, including Flickr2K~\cite{lim2017enhanced_flickr2k}, DIV2K~\cite{Agustsson_2017_div2k}, and DIV8K~\cite{gu2019div8k}, we filtered out samples with less than 5\% blur region and manually corrected detection errors of estimated blur maps. Besides, we collected 315 HQ samples from the web using the same processing, resulting in 1,362 valid samples in total.

\begin{figure*}[t]
    \centering
    \includegraphics[width=0.99\linewidth]{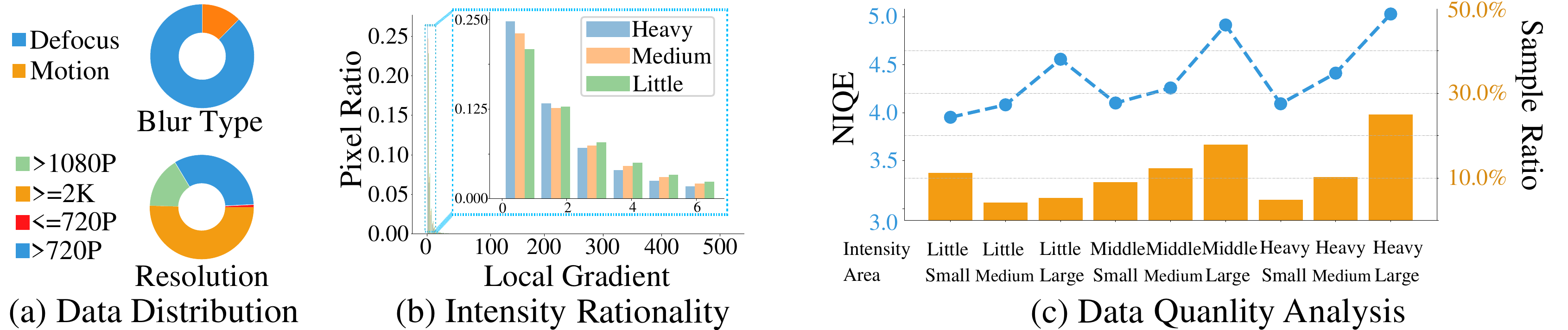}
    \caption{Data analysis of the ReBlurSR dataset. (a) The sample distribution of image resolution and blur type. (b) The local gradient in the blur region of different blur intensity subsets. (c) The sample distribution and the HR images' quality assessment based on NIQE~\cite{niqe} of different blur area sizes and blur intensities in ReBlurSR.}
\label{fig:data_analysis}
\end{figure*}

\textit{3)} 
For synthetic imagery, we developed a pipeline to create synthetic blur images using Large Language Models (LLMs) and diffusion-based methods. As shown in \cref{fig:data_analysis}b, the process begins with extracting concise image descriptions from real datasets using the GIT~\cite{wang2022git} method. GPT-3.5~\cite{OpenAI2023} diversifies these descriptions, specifying blur types. The Stable Diffusion~\cite{ldm} model transforms these descriptions into 512×512 images, which are then upscaled to 2048×2048 due to computational limitations. After blur detection and manual verification to remove errors, we produced 601 synthetic samples (more details in the Supp.).

\textit{4)} Furthermore, we applied the filtering process to real-world samples from the validation sets of DIV2K~\cite{Agustsson_2017_div2k}, CUHK~\cite{CUHK} and EBD~\cite{EBD}, resulting in the collection of 120 samples for the ReBlurSR-Test set.

In total, ReBlurSR comprises 2,210 real blur images and 601 synthetic blur images for training purposes, alongside 120 real blur images designated for testing. This collection is nearly \textbf{\textit{threefold}} larger than the widely used DIV2K and \textbf{\textit{1.14 times}} larger than Flickr2K, which is now accessible online for public use.

\noindent\textbf{Data Partition}
Following prior blur research~\cite{tang2020defusionnet,abuolaim2020defocus}, our ReBlurSR dataset encompasses the two prevalent types of blur found in natural images: 
\textbf{\textit{1)}} Defocus Blur primarily arises from aperture settings. Smaller apertures reduce it, but larger ones cause defocus blur for objects outside the focal plane, intensifying with distance from the focal plane. Modern photography often uses it to highlight the foreground by blurring less important areas. \textbf{\textit{2)}} Motion Blur results from relative movement during exposure, such as fast-moving subjects or camera shake, and is more pronounced in low-light conditions requiring longer exposures. While previous studies focused on correcting unintended blur, our work aims to preserve professionally intended motion blur.
For fine-grained assessment, we conducted a detailed dataset partition based on the following criteria:

\textbf{\textit{Blur Area Size:}} We categorized each sample based on the blur area's proportion of the total image size into three groups: small ($<$45\%), medium (45\%$\sim$55\%), and large ($>$55\%). In defocus blur samples, the blur area size predominantly falls in the small to medium range, whereas in motion blur samples, it is mostly medium to small and large. Real samples are shown in \cref{fig:data_split}b.

\textbf{\textit{Blur Intensity:}} Given the challenges in quantitatively predicting blur intensity, we relied on human visual assessment to classify images into three intensity levels: little (minimal edge and texture loss), middle (noticeable edge overlap and significant texture loss), and heavy (extensive edge loss and almost complete texture elimination), as shown in \cref{fig:data_split}b. To intuitively assess the quality of our classification, we calculated the pixel-level gradient in blur regions of different intensity subsets. As shown in \cref{fig:data_analysis}b, the gradient decreases with increasing intensity levels, indicating the degradation of textures and edges as expected.

\noindent\textbf{Data Analysis}
In \cref{fig:data_analysis}, we show the statistical distribution of resolution, blur type, size, and intensity in the ReBlurSR dataset's high-resolution (HR) images. The dataset mainly consists of high-quality images larger than 1920$\times$1080, with a higher prevalence of defocus blur compared to motion blur, likely due to the former's wider application. \cref{fig:data_analysis}b and \cref{fig:data_analysis}c indicate that samples with lower blur intensity have better NIQE metrics. Additionally, there is a correlation between blur intensity and blur area size: images with extensive blurring tend to have larger blur areas, while those with smaller blur areas exhibit less blurring.

\subsection{Perceptual-Blur-Adaptive Super-Resolution}
\label{sec:pbasr}
Given the current methods' disparate results on blur and general data, we posit that an ideal blur-adaptive BSR framework should meet the following criteria:
    \indent \textbf{- Robustness.} The framework must enhance blur image processing quality without compromising the performance of general data.

    \indent \textbf{- Efficiency.} The framework should integrate with various methods requiring minimal modifications, and incur negligible additional costs in inference.

\begin{figure}[b]
    \centering
    \includegraphics[width=0.90\linewidth]{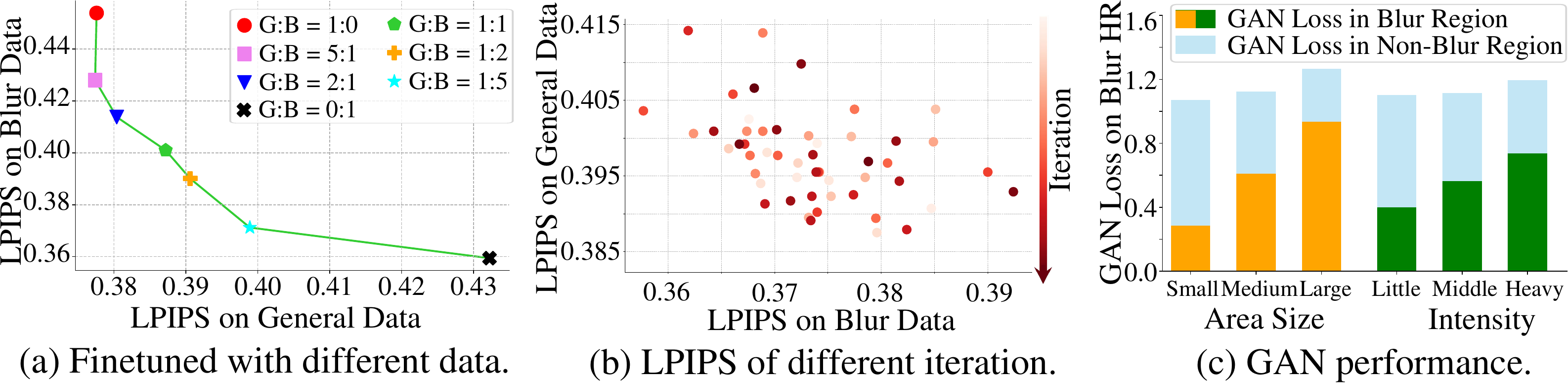}
    \caption{(a) The performance of FeMaSR~\cite{femasr_chen2022real} fine-tuned on different proportions of general (G) and blur (B) data (keeping the total training data amount the same). (b) FeMaSR's performance during late iterations which were unified trained on the combination of all general and blur images (darker colors denote later iterations). (c) The adversarial loss of general discriminator on the HR images of ReblurSR-Test.}
    \label{fig:trade_off}
\end{figure}

\noindent\textbf{Revisiting Recent Methods' Performance}
To investigate the relationship between general and blur data, and to evaluate the performance of current models, we conducted three demo experiments using FeMaSR, as detailed below:

\textbf{(1)} We explored the impact of data diversity by maintaining constant total training data amount and altering the blur data (ReBlurSR-Train) to general data ratio for FeMaSR fine-tuning. Findings in \cref{fig:trade_off}a illustrate a performance trade-off, with LPIPS shifts around 0.05 for general data and 0.08 for blur data, respectively, indicating that increasing specific data types boosts performance in its respective category but fails to simultaneously improve both.

\textbf{(2)} To investigate convergence dynamics over time, we unified both data types for an extended fine-tuning (370k iterations) and tracked FeMaSR's performance from 240k to 370k iterations. \cref{fig:trade_off}b shows LPIPS scores for general and blur data oscillating between 0.415$\sim$0.385 and 0.395$\sim$0.355, respectively, revealing a negative correlation. This further highlights the conflict between the two data types, which cannot be resolved simply by unified training.

\textbf{(3)} To explore why models excel with single-class data but falter post-merging, we analyzed adversarial loss on blur data. By comparing the average discriminator hinge loss between blur and non-blur pixels on HR Blur data, \cref{fig:trade_off}c reveals that loss in blur regions escalates from 25\%$\sim$30\% to 150\%$\sim$300\% relative to non-blur areas as blur intensity and blur area size grow, signifying diminished confidence and discriminative capacity of the classifier in blur regions.

In summary, the existing framework struggles with robustness and efficiency when handling both blur and general data simultaneously. It cannot improve performance on both data types at once, indicating a robustness issue, while splitting inference parameters by data type contradicts efficiency principles. To address these challenges, we propose the Perceptual-Blur-adaptive Super-Resolution (PBaSR) framework, which effectively merges blur and general data into a unified model without extra inference or deployment costs. The PBaSR framework incorporates the Cross Disentanglement Module (CDM) and the Cross Fusion Module (CFM), which will be detailed in the sections below.

\subsubsection{Cross Disentanglement Module}

\begin{figure*}[t]
    \centering
    \includegraphics[width=\linewidth]{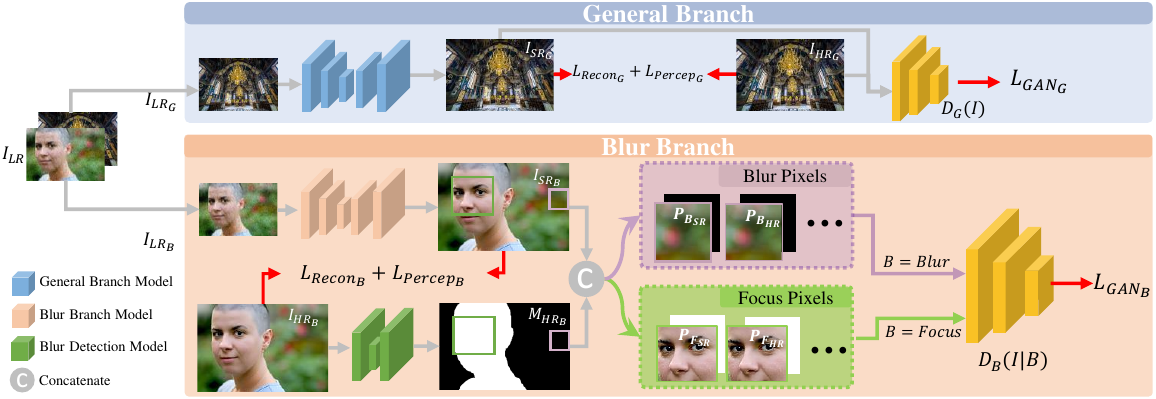}
    \caption{The structure of the Cross Disentanglement Module (CDM).}
\label{fig:CDM}
\end{figure*}

In this part, we introduce the Cross Disentanglement Module (CDM). Insights from \cref{fig:trade_off}a and \ref{fig:trade_off}b show that separate training on distinct data types enhances performance in their respective datasets. Inspired by this, we created the CDM with two branches for independently processing each data type, enabling disentanglement at the sample level for general and blur data. The structure of CDM is detailed in \cref{fig:CDM}.

\begin{figure*}[tbp]
    \centering
    \begin{minipage}{0.51\linewidth}
        \centering
        \includegraphics[width=\columnwidth]{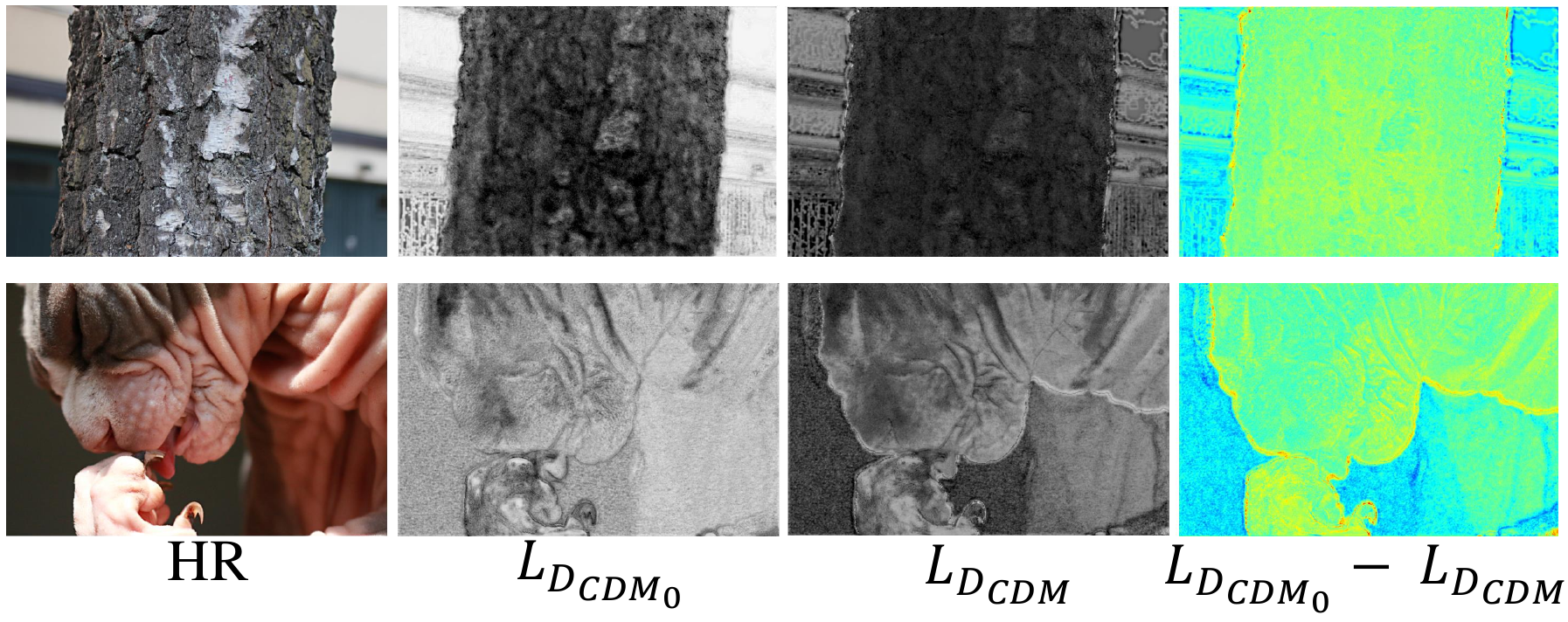}
        \caption{Map visualization of the discriminator loss in the HR blur samples. More blue for larger residual and red for the reverse.}
        \label{fig:CDM_abl}
    \end{minipage}
    \begin{minipage}{0.45\linewidth}
        \centering
        \tabcaption{Discriminator loss on HR defocus data in ReblurSR-Test. CDM$_0$ denotes CDM trained without the blur conditional adversarial loss.
             `Blur', `Focus', and `All' in $L_{D}$ on blur data denote the corresponding regions' loss.}
            \label{tab:abl_cdm}
        \resizebox{\linewidth}{!}{
        \setlength{\tabcolsep}{0.4mm}{
            \begin{tabular}{l|ll|lll}
            \hline
                \multirow{2}*{Method} & \multicolumn{2}{c|}{LPIPS} &  \multicolumn{3}{c}{$L_{D}$ on blur data} \\ \cline{2-6}
                ~ & Blur & General & Blur & Focus & All \\ \hline
                Baseline & 0.4178 & 0.3848 & 0.7047 & 0.8588 & 1.5636 \\
                CDM$_0$ & 0.3786 & \textbf{0.3809} & 0.5677 & \textbf{0.5926} & 1.1603 \\ 
                CDM & \textbf{0.3564} & 0.3826 & \textbf{0.3974} & 0.6051 & \textbf{1.0026} \\ 
                \hline
            \end{tabular}
        }
        }

        \end{minipage}
\end{figure*}

First, the general branch processes general data using existing training strategies, while the distinct blur branch is tailored for training on blur data. \cref{fig:trade_off}c indicates that training blur data within the conventional unconditional adversarial learning framework leads to suboptimal results, for which the reduced confidence of the discriminator in the blur regions is one of the primary factors. Given the 0.3$\sim$0.5 GAN loss disparity between blur and non-blur areas, we advocate for customized loss adjustments in the blur branch for better optimization in blur regions, diverging from general data strategies.
We posit that incorporating a priori information as an additional loss calculation guide is beneficial. Thus, blur maps from the ReBlurSR dataset, derived from human annotation and auto-detection via D-DFFNet with a 0.5 threshold for binary conversion, have been integrated. These maps, with a manual-to-auto annotation ratio of about 20\%:80\%, inform the redefined GAN loss for the blur branch. As shown in \cref{eq:gan_b}, the blur map $M$ is concatenated with the input image and then input into the discriminator $D_{B}$ for conditional discrimination. By incorporating the blur map as a conditional factor, the adversarial loss $L_{D_{B}}$ is computed in a manner that respects the distinct characteristics of different regions,
\begin{equation}
    \begin{aligned}
     L_{D_{B}}=\frac{1}{W H C} \times \sum_{c=1}^C \sum_{w=1}^{W} \sum_{h=1}^{H}\left\{\left(1-D_{B}(I_{HR_{B}}|M)_{w, h, c}\right)+\left( 1 + D_{B}(I_{SR_{B}}|M)_{w, h, c}\right)\right\}.
    \end{aligned}
    \label{eq:gan_b}
\end{equation}
Thus, with these adjustments, we believe that the blur branch will more adeptly and efficiently process blur data. To support this, we executed a simple experiment to assess the impact of integrating the a priori blur map condition. As indicated in \cref{tab:abl_cdm}, employing the blur condition led to a reduction in the loss of blur pixels by approximately 0.17 and yielded a 0.022 improvement in LPIPS for blur samples when compared to the baseline classifier's performance on the ReBlurSR-Test HR data, without markedly impacting the loss in non-blur regions. Furthermore, visual analysis of the discriminator loss residual map on specific samples, depicted in \cref{fig:CDM_abl}, further corroborates that the inclusion of the blur condition results in a notable decrease in the loss within blur regions. These observations affirm that the blur map serves as a valuable auxiliary for more accurate discrimination in blur regions. 

\subsubsection{Cross Fusion Module}

\begin{table}[b]
    \centering
        \caption{LPIPS  comparison of FeMaSR trained unified on both kinds of data and the averaged model of two branches of CDM$_{\mathrm{FeMaSR}}$  after being trained separately.}
    \label{tab:cdm_prob}
        \resizebox{\linewidth}{!}{
        \setlength{\tabcolsep}{3mm}{
        \centering
        
        \begin{tabular}{c|c|ccc}
            
            \hline
            \multirow{2}*{Test Data} & \multirow{2}*{Unified Trained FeMaSR} & \multicolumn{3}{c}{CDM$_{\mathrm{FeMaSR}}$ only} \\
            ~ & ~ & General Branch & Blur Branch & Poster-Average \\ \hline
            General Data &  0.3976 & 0.3855 & 0.4234 & 0.3890 \\
            Blur Data & 0.3714 & 0.4100 & 0.3649 & 0.3711 \\
            \hline
        \end{tabular}
        }
    }
    
\end{table}

\begin{figure}[btp]
    \centering
    \includegraphics[width=0.95\linewidth]{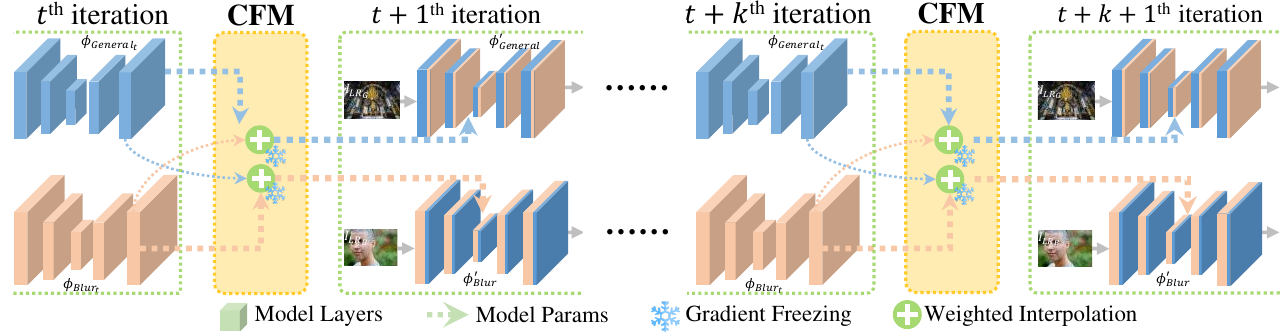}
    \caption{The structure of the Cross Fusion Module (CFM). `t' denotes the iteration number. `k' denotes the interval of CFM operation.}
    \label{fig:CFM}
\end{figure}

The CDM enables us to attain commendable results on both blur and general data. Yet, as outlined in \cref{sec:pbasr}, we aim to avoid adding complexity and computational demands during inference, a challenge given CDM's requirement for doubled parameters and extra data type distinction. An initial thought might be to average the branch weights after training, but this method falls short of expectations. \cref{tab:cdm_prob} shows that while each branch excels on its target data type, yielding about 0.01 average LPIPS improvement over the unified training baseline, the performance after weight averaging (\cref{tab:cdm_prob} col.4) is still similar to that of the unified baseline (\cref{tab:cdm_prob} col.1), indicating a persistent sub-optimal trade-off. This issue stems from inadequate communication between the branches during training. To overcome this and improve performance without additional inference costs, we propose the Cross Fusion Module (CFM), with its structure and operational phases detailed below and in \cref{fig:CFM}.

\paragraph{Training Stage}
During the training phase, our goal is for the weights from both the general and blur branches to communicate effectively. Given the need for efficiency in a multi-GPU training setup, we aim to optimize communication costs while enhancing quality. Therefore, we argue that an optimal communication strategy should meet two key criteria: \textbf{first}, to achieve optimal results on both kinds of data, the weights of the general $W_{G}$ and blur $W_{B}$ branches should exhibit a certain level of difference; \textbf{second}, to ensure effective fusion and inference, communication intervals should not be excessively long. Therefore, we adopt weight interpolation as a cost-effective method of cross-branch communication, balancing weight distance dynamically with infrequent updates. Specifically, we interpolate the weights of the general branch $W_{G}$ and the blur branch $W_{B}$ at every $k$ iteration adaptive to the cross-branch distance and apply these interpolated weights ($W'_{G, B}$) to the subsequent iteration,
\begin{equation}
    W_{G}^{'} = \lambda W_{G} + (1-\lambda)W_{B}, W_{B}^{'} = \lambda W_{B} + (1-\lambda)W_{G}, \lambda = \lambda_0 + (1 - \lambda_0)\frac{W_{G} \cdot W_{B}}{2\|W_{G}\| \|W_{B}\|}.
    \label{eq:cfm}
\end{equation}

\paragraph{Inference Stage} 
After training, we combine the weights from both branches to create a unified model. This involves equal interpolation of both branches to establish the final model parameters, expressed as $W_{PBaSR} = W_{G}/2 + W_{B}/2$.

\begin{figure}[b]
    \centering
    \includegraphics[width=0.92\linewidth]{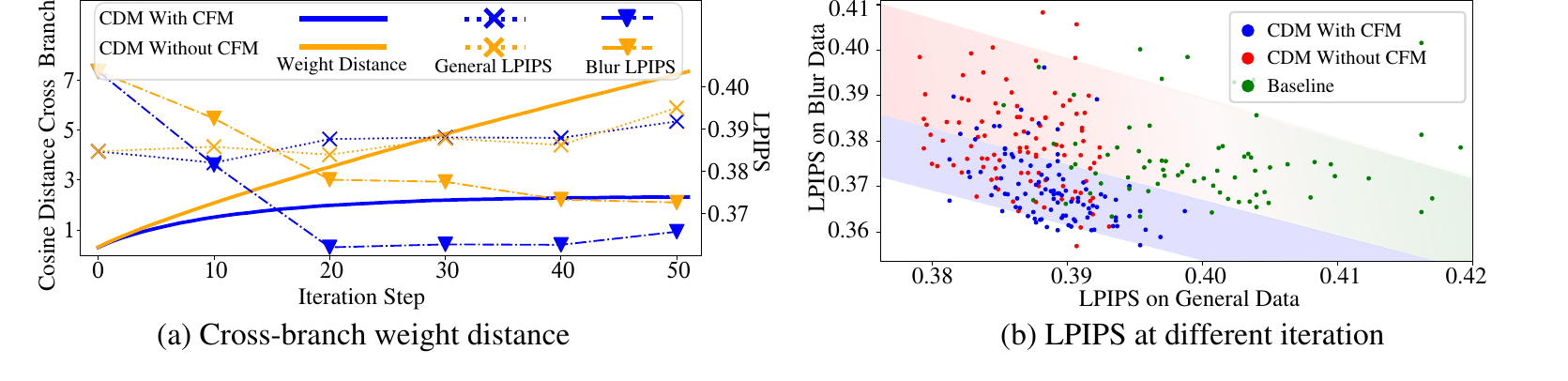}
    \caption{Comparison of CDM trained with and without CFM. (a) The cosine distance between two branches' weights at different iterations. (b) The LPIPS on general and blur data at different iterations. `Baseline' denotes the unified training as in \cref{fig:trade_off}b.}
    \label{fig:cfm_abl}
\end{figure}
The CFM and CDM do not modify the model's architecture, ensuring the inference process aligns with the original model and avoids extra inference or deployment costs. CFM's role in maintaining weight distance between different branches during training allows for the effective fusion of priors for different data types via simple interpolation averaging during inference. For clarity, we conducted an ablation analysis comparing the CDM trained with and without CFM. As shown in \cref{fig:cfm_abl}a, the cosine distance between the two branches' weights is effectively constrained within a certain range by CFM, while the distance without CFM fluctuates obviously. Besides, \cref{fig:cfm_abl}b shows that the LPIPS of CDM without CFM on general and blur data are entangled in the same region as the unified trained baseline, whereas CFM achieves a 0.01-0.015 LPIPS improvement.
% !TEX root = main.tex

\section{Experiment}

\subsection{Experimental Setup}

\noindent\textbf{Dataset}

\noindent\textit{Training Datasets:} For general training data, we employed widely recognized super-resolution datasets, including DIV2K~\cite{Agustsson_2017_div2k}, Flickr2K~\cite{lim2017enhanced_flickr2k}, and DIV8K~\cite{gu2019div8k}, covering a broad spectrum of high-resolution, real-world images. Regarding blur data, we utilized the ReBlurSR dataset as the primary source. We applied the degradation model from BSRGAN~\cite{bsrgan_zhang2021designing} for low-resolution (LR) image synthesis.

\noindent\textit{Testing Datasets:}
We assessed performance on general data using benchmark datasets including DIV2K~\cite{Agustsson_2017_div2k}, Urban100~\cite{urban100}, BSDS100~\cite{BSDS100}, Manga109~\cite{manga109}, Set14~\cite{Set14}, and Set5~\cite{Set5}. For blur data, we evaluated methods using the ReBlurSR-Test dataset. LR inputs are generated from a mixed degradation of BSRGAN~\cite{bsrgan_zhang2021designing} and RealESRGAN~\cite{wang2018esrgan}. Considering experimental efficiency, DIV2K~\cite{Agustsson_2017_div2k} is used as the default general benchmark in all ablation analyses.\newline

\noindent\textbf{Evaluation}
    We use perceptual metrics as the main performance evaluation metrics. Specifically, we employ six widely recognized metrics to evaluate the performance of our proposed method, namely Learned Perceptual Image Patch Similarity (LPIPS)~\cite{LPIPS_zhang2018unreasonable}, Visual Information Fidelity (VIF)~\cite{sheikh2006image}, Gradient Magnitude Similarity Deviation (GMSD)~\cite{xue2013gradient}, Visual Saliency Index (VSI)~\cite{vsi_6873260}, Deep Image Structure and Texture Similarity (DISTS)~\cite{ding2020iqa}, and Attention-based Hybrid Image Quality (AHIQ) assessment~\cite{lao2022attentions}. For reference, the PSNR/SSIM and more perceptual metrics~\cite{ckdn, ghildyal2022stlpips,chen2023topiq} results are also provided in the supplement.\newline

\noindent\textbf{Implementation Details}
    In the implementation, branch initial weights are based on those from the general BSR task, using officially released weights. The optimization process for each branch is conducted independently, utilizing an Adam optimizer with a learning rate set at $1\times10^{-4}$. We set the batch size of each GPU and the patch size of HR to 16 and 256, respectively. Each batch is composed of an equal amount of general and blur data. The $\lambda_0$ and $k$ in CFM are set to 0.99 and 20, respectively.  The training spanned a total of 200,000 iterations, executed on two NVIDIA V100 GPUs, with PyTorch serving as the programming framework. All experiments are conducted using x4 upscaling.

    \begin{table*}[tbp]
        \renewcommand{\arraystretch}{1.}
    \centering
    \caption{Quantitative comparison of PBaSR with the SOTA methods on blur data and general data. The blur validation set is ReBlurSR-Test. The general validation set consists of six benchmarks, including DIV2K, Urban100, BSDS100, Manga109, Set14, and Set5. The \best{best} and \second{second-best} results are bolded in black and red, respectively.}
    \label{tab:sota}
    \vspace{-0.2cm}
    \resizebox{\linewidth}{!}{
    \setlength{\tabcolsep}{0.8mm}{
        \large
    \begin{tabular}{c|c|ccccccc|ccc}
        
        \hline
        \multirow{2}*{Data} & \multirow{2}*{Metric} & \multirow{2}*{\makecell[c]{SwinIR \\ (2021)}} & \multirow{2}*{\makecell[c]{Real-ESRGAN \\ (2021)}} & \multirow{2}*{\makecell[c]{MM-RealSR \\ (2022)}} & \multirow{2}*{\makecell[c]{FeMaSR  \\ (2022)}}& \multirow{2}*{\makecell[c]{CAL-GAN \\ (2023)}} & \multirow{2}*{\makecell[c]{HAT \\ (2023)}} &\multirow{2}*{\makecell[c]{SRFormer \\ (2023)}} & \multirow{2}*{${\mathrm{\networkname_{ESRGAN}}}$} & \multirow{2}*{${\mathrm{\networkname_{FeMaSR}}}$} & \multirow{2}*{${\mathrm{\networkname_{SRFormer}}}$} \\
        ~ & ~ &~ &~ &~ &~ &~ &~ &~ &~ &~ &~ \\ \hline 
    
        \multirow{6}*{\makecell[c]{Defocus \\ Blur}} & LPIPS $\downarrow$     & 0.4048 & 0.4199 & 0.4270 & 0.4037 & 0.4511 & 0.3924  & 0.3974 & 0.3986 & \best{0.3564} & \second{0.3740} \\
        ~ & AHIQ $\uparrow$                                                   & 0.2396 & 0.2215 & 0.2204 & 0.2323 & 0.1857 & 0.2282  & 0.2394 & 0.2386 & \best{0.2670} & \second{0.2620} \\
        ~ & DISTS $\downarrow$                                                & 0.2074 & 0.2313 & 0.2342 & 0.1861 & 0.2607 & 0.2502  & 0.2139 & 0.1952 & \best{0.1733} & \second{0.1734} \\
        ~ & VIF $\uparrow$                                                    & 0.0873 & 0.0842 & 0.0833 & 0.0862 & 0.0821 & 0.0859  & 0.0896 & 0.0907 & \best{0.0938} & \second{0.0931} \\
        ~ & GMSD $\downarrow$                                                 & 0.1843 & 0.1900 & 0.1851 & 0.1781 & 0.1854 & 0.1888  & 0.1796 & 0.1773 & \second{0.1737} & \best{0.1709} \\
        ~ & VSI $\uparrow$                                                    & 0.9565 & 0.9547 & 0.9552 & 0.9602 & 0.9573 & 0.9528  & 0.9577 & 0.9609 & \second{0.9621} & \best{0.9625} \\\hline
        \multirow{6}*{\makecell[c]{Motion \\ Blur}} & LPIPS                   & 0.4252 & 0.4104 & 0.4238 & 0.4659 & 0.4605 & 0.3856 & 0.4145 & \second{0.3791} & \best{0.3624} & 0.3887 \\ 
        ~ & AHIQ $\uparrow$                                                   & 0.2852 & 0.2920 & 0.2771 & 0.2330 & 0.2343 & \second{0.2964} & 0.2846 & \best{0.2986} & 0.2884 & 0.2931 \\ 
        ~ & DISTS $\downarrow$                                                & 0.2105 & 0.2212 & 0.2324 & 0.2097 & 0.2415 & 0.2250 & 0.2127 & 0.1907 & \best{0.1771} & \second{0.1844} \\
        ~ & VIF $\uparrow$                                                    & 0.1192 & 0.1135 & 0.1125 & 0.1225 & 0.1121 & 0.1171 & 0.1238 & 0.1244 & \best{0.1320} & \second{0.1292} \\ 
        ~ & GMSD $\downarrow$                                                 & 0.1713 & 0.1731 & 0.1706 & 0.1775 & 0.1770 & 0.1679 & 0.1670 & \second{0.1648} & \best{0.1629} & 0.1649 \\ 
        ~ & VSI $\uparrow$                                                    & 0.9744 & 0.9733 & 0.9721 & 0.9760 & 0.9749 & 0.9723 & 0.9748 & 0.9771 & \best{0.9782} & \second{0.9781} \\ \hline
        \multirow{6}*{General}& LPIPS $\downarrow$                            & 0.4202 & 0.4703 & 0.4952 & 0.3986 & 0.4775 & 0.4850 & 0.4284 & 0.4463 & \best{0.3912} & \second{0.3937} \\
        ~ & AHIQ $\uparrow$                                                   & 0.1944 & 0.1890 & 0.1707 & 0.1932 & 0.1557 & 0.1828 & 0.1990 & 0.1860 & \second{0.2190} & \best{0.2223} \\
        ~ & DISTS $\downarrow$                                                & 0.2475 & 0.2839 & 0.3060 & \second{0.2303} & 0.2928 & 0.3115 & 0.2615 & 0.2587 & 0.2331 & \best{0.2229} \\
        ~ & VIF $\uparrow$                                                    & 0.0646 & 0.0568 & 0.0563 & 0.0619 & 0.0560 & 0.0598 & \best{0.0674} & 0.0617 & 0.0668 & \second{0.0673} \\
        ~ & GMSD $\downarrow$                                                 & 0.2387 & 0.2515 & 0.2536 & \best{0.2225} & 0.2418 & 0.2523 & 0.2402 & 0.2350 & 0.2261 & \second{0.2243} \\
        ~ & VSI $\uparrow$                                                    & 0.9176 & 0.9080 & 0.9059 & 0.9257 & 0.9139 & 0.9060 & 0.9174 & 0.9192 & \second{0.9257} & \best{0.9260} \\
        \hline
    \end{tabular}
    }
    }
\vspace{-0.3cm}
    \end{table*}

\subsection{Comparison with State-of-the-art Methods}
% \vspace{-0.1cm}
For comprehensive and reasonable comparative analysis, 
we conducted our PBaSR on three widely recognized BSR methods, namely Real-ESRGAN~\cite{realesrgan_wang2021real}, FeMaSR~\cite{femasr_chen2022real}, and SRFormer~\cite{zhou2023srformer}. We compared them with four other state-of-the-art (SOTA) BSR methods (including SwinIR~\cite{liang2021swinir}, MM-RealSR~\cite{mou2022mmrealsr}, CAL-GAN~\cite{park2023content}, and HAT~\cite{chen2023activating}), evaluating performance on both general and blur data. 
To ensure fairness in the comparison, 
the results for these SOTA methods were derived using the code and weights available from their official repositories. 
Additionally,
we included comparative experiments to assess the impact of fine-tuning with the ReBlurSR dataset in our ablation studies, thereby providing a more nuanced understanding of our framework's performance under varied training conditions.     
The quantitative and qualitative results are shown in \cref{tab:sota} and \cref{fig:qualitative}.

As shown in \cref{tab:sota}, our $\mathrm{PBaSR_{FeMaSR}}$ (\cref{tab:sota} col.9) outperforms the leading method for blur data, HAT~\cite{chen2023activating} (\cref{tab:sota} col.6), with an improvement of 0.04$\sim$0.09 in LPIPS~\cite{LPIPS_zhang2018unreasonable}, while maintaining comparable results on general data (\cref{tab:sota} last 6 rows) with the best-performing FeMaSR~\cite{femasr_chen2022real} (\cref{tab:sota} col.4) and SRFormer~\cite{zhou2023srformer} (\cref{tab:sota} col.7) across all metrics. \cref{fig:qualitative} presents several examples showcasing different types of blur. Taking the first sample in \cref{fig:qualitative}, most general BSR methods (\cref{fig:qualitative} cols. 1, 2, and 4.) generate artifacts and over-texturization in blur regions. In contrast, our PBaSR (\cref{fig:qualitative} col.5) exhibits superior blur preservation both visually and quantitatively. While CAL-GAN~\cite{park2023content} (\cref{fig:qualitative} col.3 row.1) preserves blur relatively well, it tends to over-smooth and lose texture in focused areas (\cref{fig:qualitative} col.3, row.2). PBaSR, however, not only enhances restoration in blur regions (\cref{fig:qualitative} row.1 and 3) but also achieves the best perception and LPIPS metrics in focused areas (\cref{fig:qualitative} Row.2 and 4). In summary, both quantitative and qualitative analyses reveal that PBaSR effectively alleviates recent methods' limitations in blur image blind super-resolution and also maintains comparative texture restoration performance in general blind super-resolution tasks.

    \begin{figure*}[tbp]
        \centering
        \includegraphics[width=\linewidth]{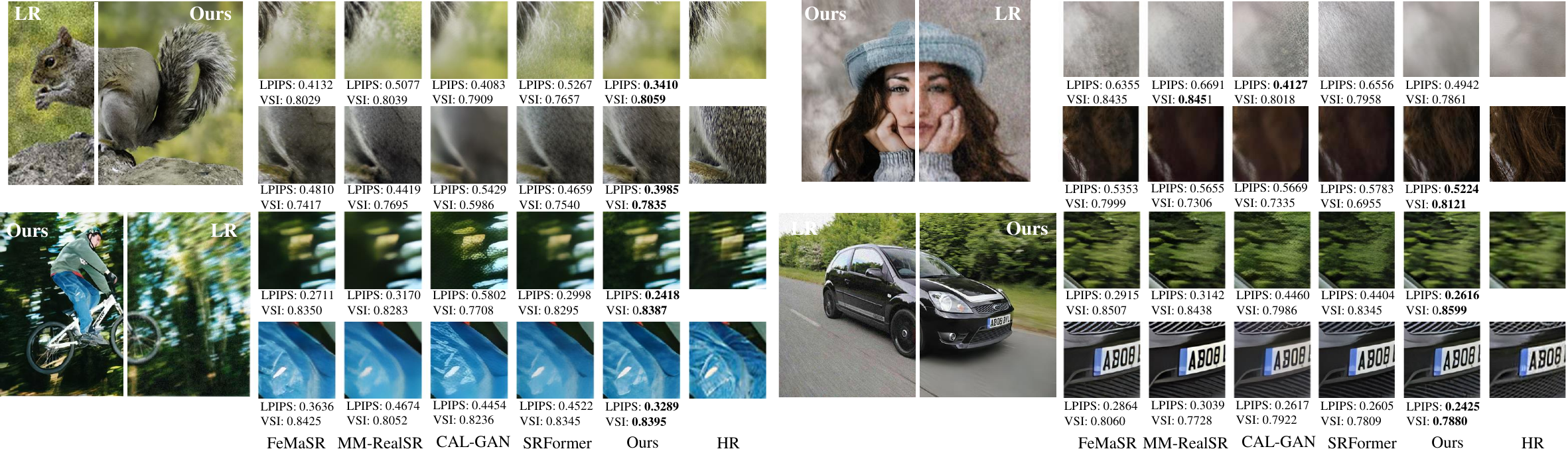}
        \caption{Comparison of ${\mathrm{PBaSR_{FeMaSR}}}$ with SOTA methods on blur regions (up) and non-blur regions (down). More results are provided in the supplement.}
        \label{fig:qualitative}
    \end{figure*}

\subsection{Ablation Study}

\noindent\textbf{Scalability for Blur Data}
To assess PBaSR's flexibility with different blur types, we categorized the ReBlurSR-Train set by blur types and progressively included various blur types during training, beginning with defocus and then motion blur data. We conducted detailed evaluations on subsets of distinct blurs in the ReBlurSR-Test. \cref{tab:blur_type} shows that adding defocus data (from the 1st to 2nd and the 4th to 5th rows) enhances performance on the defocus blur test for a 0.04 LPIPS improvement without a heavy negative impact on general data performance. Likewise, incorporating all blur data (from the 2nd to 3rd and the 5th to 6th rows) leads to enhancements across all blur data. These findings reveal PBaSR's robust scalability across different blur types, effectively adjusting to new blur categories while maintaining performance on previously trained types.

\begin{figure*}[bp]
    \centering
    \begin{minipage}{0.52\linewidth}
    \large    
    \renewcommand{\arraystretch}{1.26}
    \centering
    \tabcaption{Scalability of $\mathrm{PBaSR}_{\mathrm{FeMaSR}}$ on different blur subsets. ``+G'', ``+D'', and ``+M'' denote general, defocus blur, and motion blur data, respectively. Top: Blur Intensity; Bottom: Blur area size.}
    \resizebox{\linewidth}{!}{
    \setlength{\tabcolsep}{0.3mm}{
    \begin{tabular}{ccc|cccc|cccc|c}
    \hline
        \multicolumn{3}{c|}{Train Data} & \multicolumn{9}{c}{LPIPS} \\ \hline
        +G & +D & +M & \multicolumn{4}{c|}{Defocus Blur} & \multicolumn{4}{c|}{Motion Blur} & \multirow{2}{*}{General} \\ \cline{1-11}
        ~ & ~ & ~ & Little & Middle & Heavy & All & Little & Middle & Heavy & All & ~ \\ \hline
        $\checkmark$ & $\times$ & $\times$ & 0.4237 & 0.4014 & 0.3689 & 0.4037 & 0.4871 & 0.4706 & 0.4461 & 0.4659 & 0.3848 \\
        $\checkmark$ & $\checkmark$ & $\times$ & 0.3809 & 0.3717 & 0.3315 & 0.3657 & 0.4250 & 0.3833 & 0.3439 & 0.3800 & 0.3856 \\
        $\checkmark$ & $\checkmark$ & $\checkmark$ & \best{0.3764} & \best{0.3595} & \best{0.3190}  & \best{0.3564} & \best{0.3995} & \best{0.3782} & \best{0.3205} & \best{0.3624} & \best{0.3826} \\ \hline
        ~ & ~ & ~ & Small & Medium & Large & All & Small & Medium & Large & All & ~ \\ \hline
        $\checkmark$ & $\times$ & $\times$ & 0.4178 & 0.4180 & 0.3368 & 0.4037 & 0.4352 & 0.3898 & 0.4765 & 0.4659 & 0.3848 \\
        $\checkmark$ & $\checkmark$ & $\times$ & 0.3719 & 0.3848 & 0.3043 & 0.3657 & 0.3691 & 0.3323 & 0.3850 & 0.3800   & 0.3856 \\
        $\checkmark$ & $\checkmark$ & $\checkmark$ & \best{0.3660} & \best{0.3774} & \best{0.2901} & \best{0.3564} & \best{0.3559} & \best{0.3239} & \best{0.3660} & \best{0.3624} & \best{0.3826} \\ \hline
    \end{tabular}
    }
    }
    
    \label{tab:blur_type}
    \end{minipage}
    \begin{minipage}{0.47\linewidth}
        \renewcommand{\arraystretch}{1.06}
        \centering
        \tabcaption{Generalizability across various model structures. ``Base'' denotes the official weights. ``Flop'' denotes the flop operation number when processing a $256\times256$ patch.}
        \resizebox{\linewidth}{!}{
        \setlength{\tabcolsep}{0.1mm}{
        \begin{tabular}{c|c|c|c|c|c|c}
        \hline
        \multirow{2}*{Method} & \multirow{2}*{Model Type} & \multirow{2}*{Framework} & \multirow{2}*{Flops} & \multicolumn{3}{c}{LPIPS}   \\ \cline{5-7}
            ~ & ~ & ~ & ~& Defocus & Motion & General  \\ \hline
            \multirow{3}*{ESRGAN} & \multirow{3}*{CNN} & Base & \multirow{3}*{11.7T} & 0.4199 & 0.4104 & 0.4739 \\  
            ~ & ~ & Unified Finetune & ~ & 0.4148 & 0.3910 & 0.4419 \\ 
            ~ & ~ & \networkname (ours) & ~ & \best{0.3986} & \best{0.3791} & \best{0.4390}\\ \hline
            \multirow{3}*{FeMaSR} & \multirow{3}*{VQVAE} & Base & \multirow{3}*{15.0T} & 0.4037 & 0.4586 & 0.3848 \\ 
            ~ & ~ & Unified Finetune & ~ & 0.3714 & 0.3905 & 0.3976 \\ 
            ~ & ~ & \networkname (ours) & ~ & \best{0.3564} & \best{0.3663} & \best{0.3826} \\ \hline
            \multirow{3}*{SRFormer} & \multirow{3}*{Transformer} & Base & \multirow{3}*{2.0T}  & 0.3974 & 0.4145 & 0.4170 \\ 
            ~ & ~ & Unified Finetune & ~ & 0.4139 & 0.4144 & 0.4150\\ 
            ~ & ~ & \networkname (ours) & ~ & \best{0.3740} & \best{0.3887} & \best{0.3892} \\ \hline
        \end{tabular}
        }
        }

        \label{tab:generalizability}
    \end{minipage}
\end{figure*}

\noindent\textbf{Effect of Training Data}
\begin{figure}[tbp]
    \centering
    \includegraphics[width=0.99\linewidth]{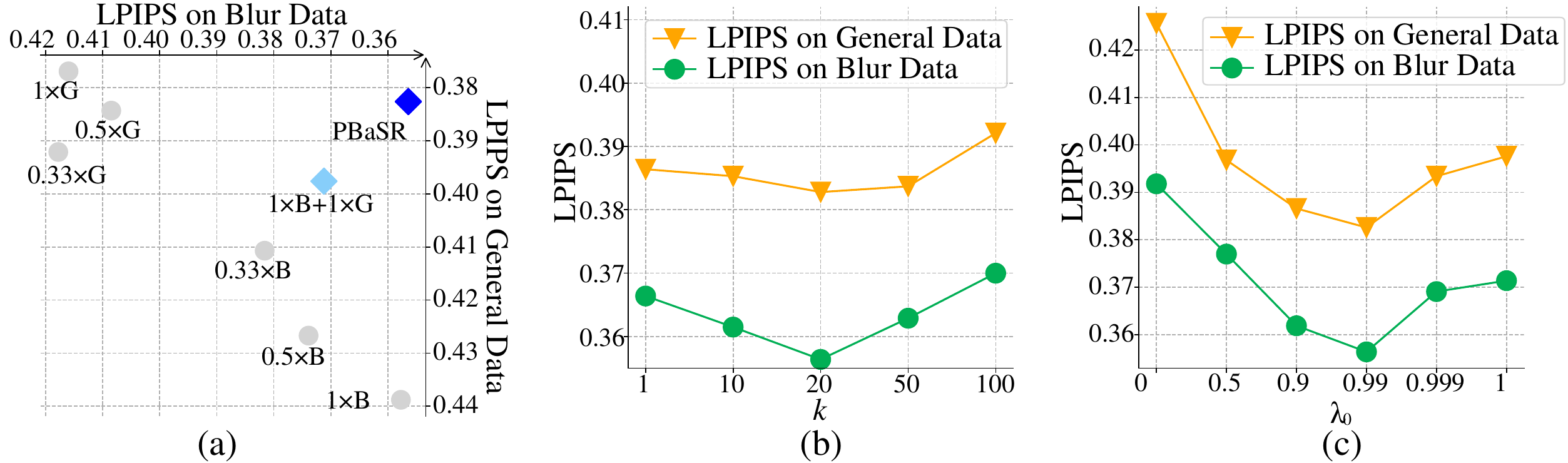}
    \vspace{-0.2cm}
    \caption{Ablations on ReBlurSR-Test (Defocus) and DIV2K-Val. (a) FeMaSR trained with different training data settings. ``$\times$N'' denotes the used ratio of the data type. ``1$\times$B + 1$\times$G'' denotes the unified training with all training data. (b)/(c) Effect of the communication frequency `$k$' and ratio interpolation `$\lambda_0$' of CFM in Eq.~\ref{eq:cfm}. 
    }
    \label{fig:data_abl}
\end{figure}
To elucidate PBaSR's effectiveness, we undertook an extensive evaluation based on FeMaSR, examining how different types and amounts of training data affect performance. \cref{fig:data_abl}a reveals that without PBaSR (gray points), training solely on general or blur data improves results for that specific category but leads to reduced performance for the other. Merging both data types for training fails to yield simultaneous gains in both areas (sky blue point). Conversely, PBaSR (blue point) shows notable improvements for both data types, evidencing the successful disentanglement and integration of these categories by the CDM and CFM.

\noindent\textbf{Generalizability across Anchor Method Structures}
To evaluate PBaSR's versatility across different SR model architectures, we implemented the PBaSR framework on three widely recognized architectures: CNN, VQVAE, and Transformers. \cref{tab:generalizability} shows that PBaSR yielded noticeable enhancements across these mainstream SR architectures. Additionally, our framework proved effective across models of varying sizes and types, from SRFormer to FeMaSR, showing consistent performance improvements regardless of model scale. Notably, SRFormer and FeMaSR saw higher gains, possibly due to their superior fitting capabilities.

\noindent\textbf{Effect of CFM Communication Strategy}
To evaluate the impact of the communication strategy in CFM, we performed ablation studies on its frequency and ratio of cross-branch interpolation with $\mathrm{PBaSR}_{\mathrm{FeMaSR}}$. Fig.\ref{fig:data_abl}b shows that either too frequent or infrequent communication ($k=1\ \mathrm{or}\ k=100$) decreases LPIPS by 0.005$\sim$0.01 due to inappropriate distances between branch weights, leading to poor interpolation results. \cref{fig:data_abl}c indicates that the absence of communication ($\lambda_0 = 1$) or overly aggressive interpolation ($\lambda_0 = 0$)  negatively affects performance while a $\lambda_0$ value between $0.9\sim0.99$ ensures performance stability. Additionally, we compared CFM against other feature communication methods like feature distillation and teacher-student learning in supplement, which also indicates the effectiveness and efficiency of CFM.

% !TEX root = main.tex

\section{Conclusion}
\label{sec:conclusion}
In this work, we explored blind super-resolution for real-world blur images and created the ReBlurSR dataset containing 2931 diverse blur images. We proposed a novel Perceptual-Blur-adaptive Super-Resolution (PBaSR) framework to address the limitations of current methods in processing blur data. Extensive evaluations on various benchmarks reveal that PBaSR significantly improves performance on real-world blur images while maintaining strong results on general data, without incurring extra inference or deployment costs.
\section*{Acknowledgements}
This work was supported by the National Natural Science Foundation of China under Grant 62072271.

% \clearpage
\nocite{*}
\bibliographystyle{splncs04}
\bibliography{main}

\clearpage
\appendix
\section{Appendix}

\subsection{Blur in General BSR Scenarios}
\begin{figure}[htbp]
    \centering
    % \vspace{-1cm}
    \includegraphics[width=0.62\linewidth]{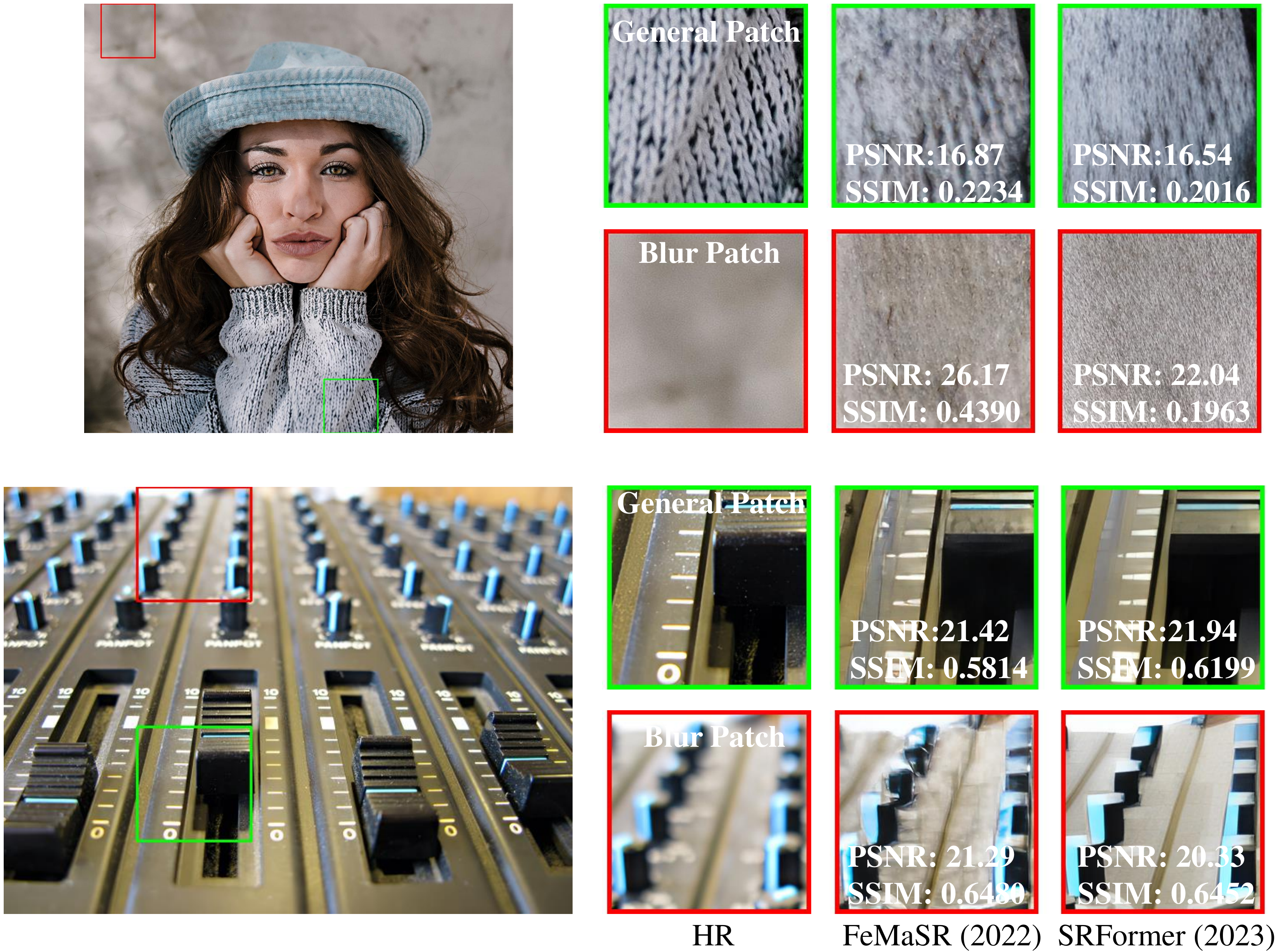}
    % \vspace{-0.4cm}
    \caption{More visual samples of the existing methods on real-world blur data.}
    \label{fig:qualitative_4}
\end{figure}

\begin{figure}[htbp]
    \centering
    \includegraphics[width=0.95\linewidth]{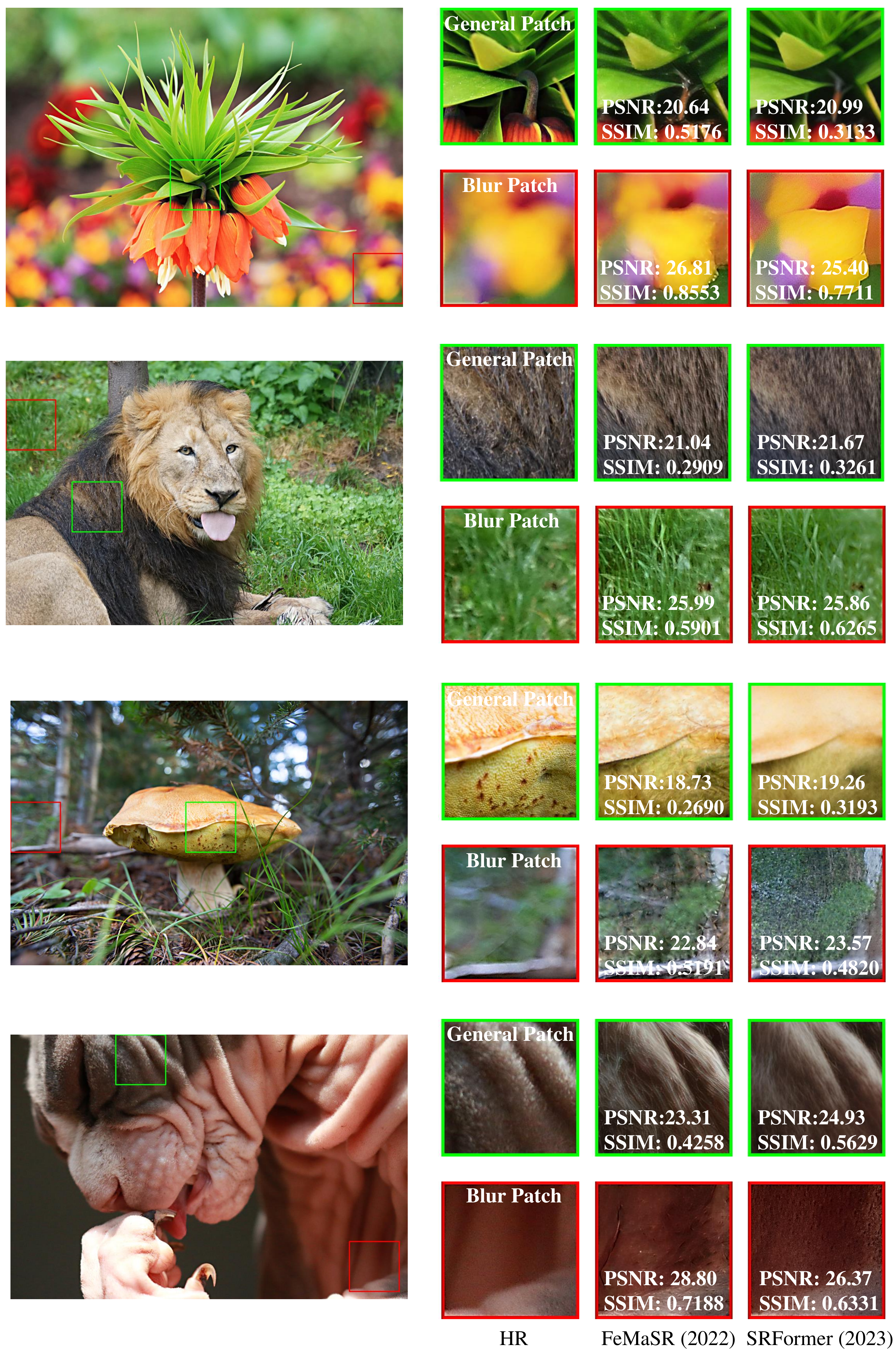}
    \caption{More visual samples of the existing methods on real-world blur data.}
    \label{fig:qualitative_3}
\end{figure}

To further illustrate the prevalence of blur in high-quality real-world images, we provide additional qualitative and quantitative phenomena as support. Qualitatively, we showcase more examples with blur in real-world data from public datasets where existing methods struggle, highlighting the wide usage of various intentional blurring techniques in real-world image scenarios and the challenges faced by recent general BSR methods, as seen in \cref{fig:qualitative_4} and \cref{fig:qualitative_3}. For instance, the images on the left side of \cref{fig:qualitative_4} show how background defocus and lens depth of field enhance subject focus and picture depth. Yet, current BSR methods, while improving texture in foreground areas (\cref{fig:qualitative_4},~\ref{fig:qualitative_3}, right side, odd row), often mismanage blurred areas with incorrect texturization and oversharpening (\cref{fig:qualitative_4},~\ref{fig:qualitative_3}, right side, even rows), diminishing the intended blur effect and overall image perception. This issue has not yet received enough attention despite advancements achieved by recent methods in general image processing.
From a quantitative perspective, we analyzed the prevalence of blur in key super-resolution benchmarks, including DIV2K, Flickr2K, and DIV8K, with findings summarized in \cref{tab:abl_dataset}. Approximately 20\% of high-resolution images in these datasets exhibit blur, further indicating its widespread presence and underscoring the importance of incorporating blur handling into super-resolution models.
\begin{table*}[bp]
    \caption{Blur proportion in commonly used super-resolution benchmark datasets.}
    \centering
    \resizebox{\linewidth}{!}{
    \setlength{\tabcolsep}{3mm}{
    \centering
    \begin{tabular}{c|cccc}
    \hline
        Dataset & DIV2K~\cite{Agustsson_2017_div2k} & Flickr2K~\cite{lim2017enhanced_flickr2k} & DIV8K~\cite{gu2019div8k} & All \\ \hline
        Blur Amount/Dataset Size & 175+22/800+100 & 606/2560 & 266/1500 & 1069/4960 \\ 
        Blur Proportion & 21.86\%+22.00\% & 23.67\% & 17.73\% & 21.55\% \\ \hline
    \end{tabular}
    }
    }
    \label{tab:abl_dataset}
\end{table*}


\clearpage
\subsection{Effectiveness of Data Synthesis}
\begin{figure}
    \centering
    \includegraphics[width=\linewidth]{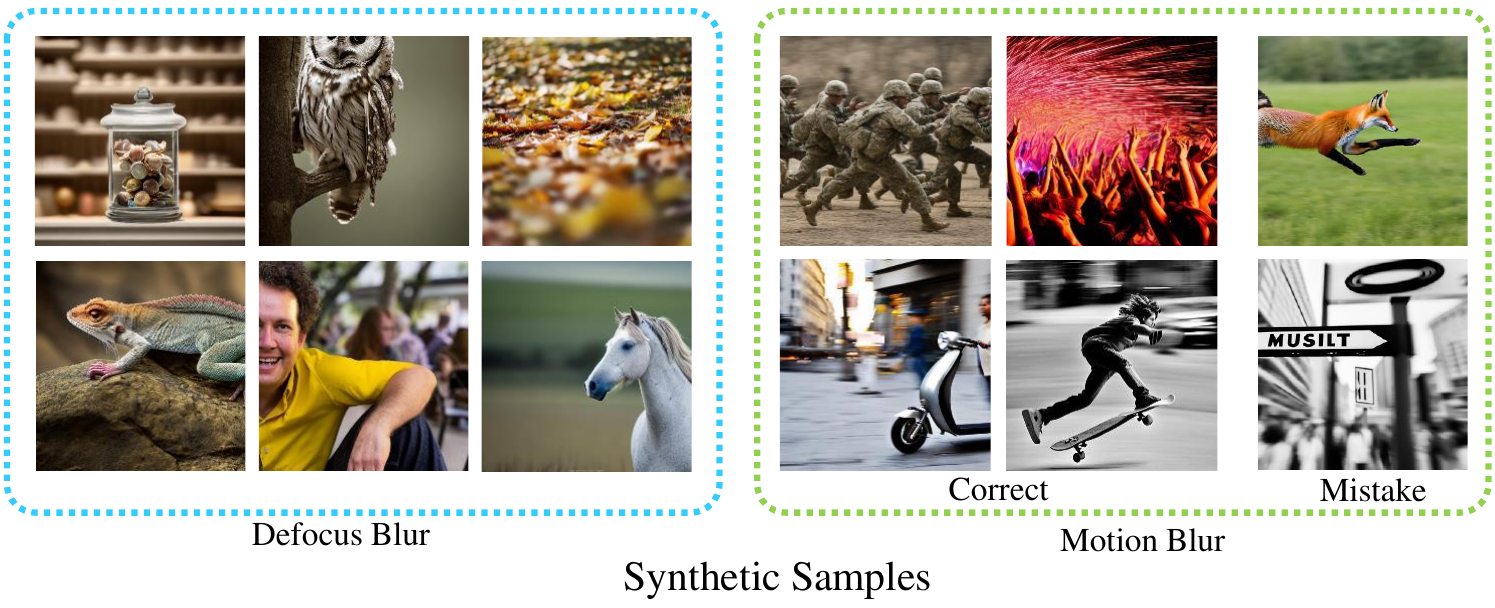}
    % \vspace{-0.3cm}
    \caption{Samples of synthetic data with different kinds of blur.}
    % \vspace{-0.3cm}
    \label{fig:syn_samples}
    % \vspace{-1cm}
\end{figure}

\begin{figure*}[thbp]
    \centering
    \begin{minipage}{0.49\linewidth}
        \centering
        % \vspace{0pt}
        \tabcaption{Effectiveness of synthetic data on the blur BSR performance of finetuned FeMaSR (without using general data). `Real' denotes only the real part of the ReblurSR-Train dataset (old 1804 samples).}
            %  \vspace{-0.2cm}
            \label{tab:abl_cdm}
        \resizebox{\linewidth}{!}{
        \setlength{\tabcolsep}{1mm}{
            \begin{tabular}{l|ll}
                \hline
                    \multirow{2}*{Finetuning Data} & \multicolumn{2}{c}{LPIPS}  \\ \cline{2-3}
                    ~ & Blur & General \\ \hline
                    ReBlurSR-Train(Real) & 0.3638 & \textbf{0.4324}  \\
                    ReblurSR-Train & \textbf{0.3508} & 0.4388  \\ 
                    \hline
                \end{tabular}
        }
        }
        \label{tab:abl_sync}
            
        \end{minipage}
    \begin{minipage}{0.49\linewidth}
        \centering
        % \vspace{0pt}
        \includegraphics[width=\columnwidth]{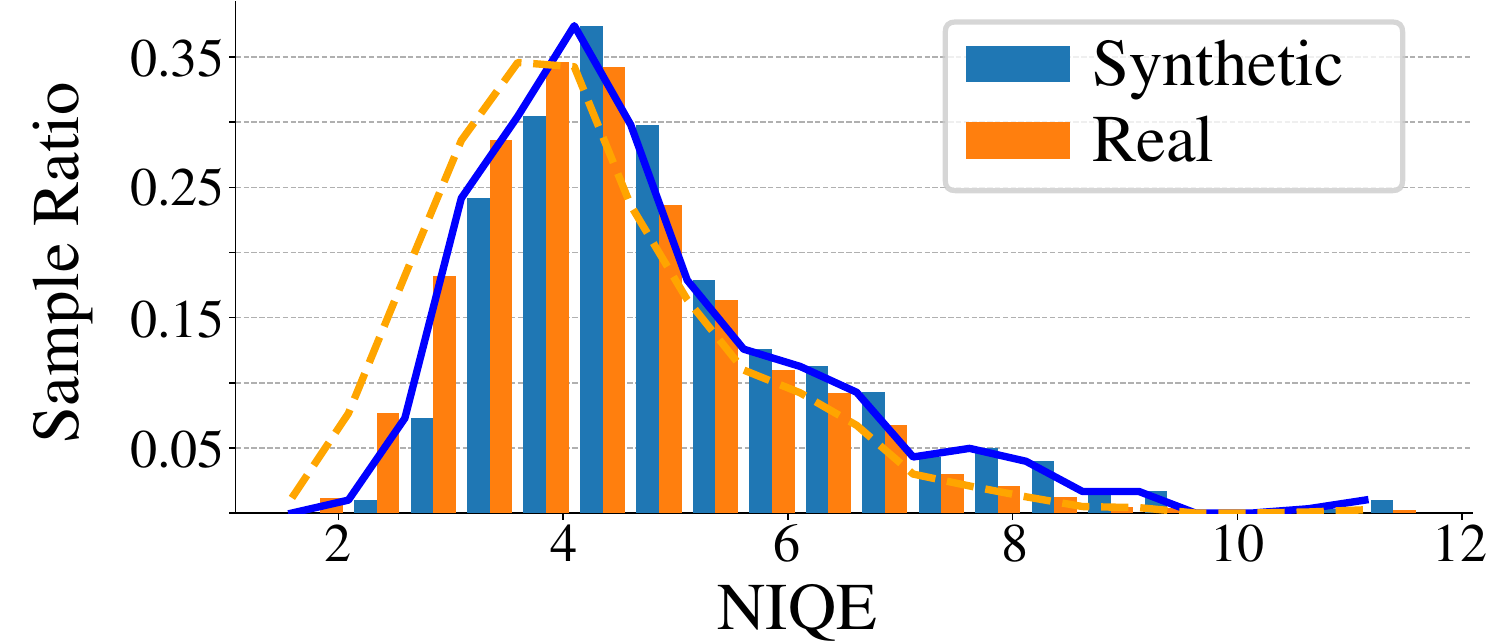}
        % \vspace{-0.7cm}
        \caption{NIQE distribution comparison between synthetic and real images in the ReBlurSR-Train dataset.}
        \label{fig:niqe}
    \end{minipage}
    
\end{figure*}

At first, we synthesized 1000 images that included both defocus and motion blur. However, we found that stable diffusion's understanding of motion blur was more confusing than its effective understanding of defocus blur, and motion blur is often misinterpreted and generated as a defocus blur-like morphology. Therefore, considering the effectiveness of training, we excluded the generated incorrect motion blur from the valid data by manual filtering and supplemented it with a corresponding number of images from online resources.
Here, we thoroughly evaluated the quality of the synthesized images from the perspectives of visual effects and quantitative metrics. From a qualitative standpoint, we present a series of synthesized images featuring defocus and motion blur in \cref{fig:syn_samples}. Quantitatively, we employed the NIQE~\cite{niqe} as a metric to assess the quality of high-resolution images, comparing the NIQE scores of both real and synthesized samples in ReBlurSR-Train in \cref{fig:niqe}. The results indicate that the synthetic data exhibits a distribution pattern akin to that of the real data, thereby affirming our process's capability to produce high-quality images that closely mimic the distribution characteristics of authentic data.
Additionally, to further validate the effect of synthetic data on model training, we compared the performance of FeMaSR fine-tuned using all data in ReBlurSR-Train and using only the real parts. As shown in \cref{tab:abl_sync}, the addition of synthetic data improves the model's LPIPS performance on blur data by about 0.013. This indicates that the synthetic data indeed expands the diversity of the training data and improves the generalization ability.


\subsection{Effectiveness of Cross Fusion Module}
\begin{figure*}[htbp]
    \centering
\begin{minipage}{0.49\linewidth}
    \centering
    \includegraphics[width=\linewidth]{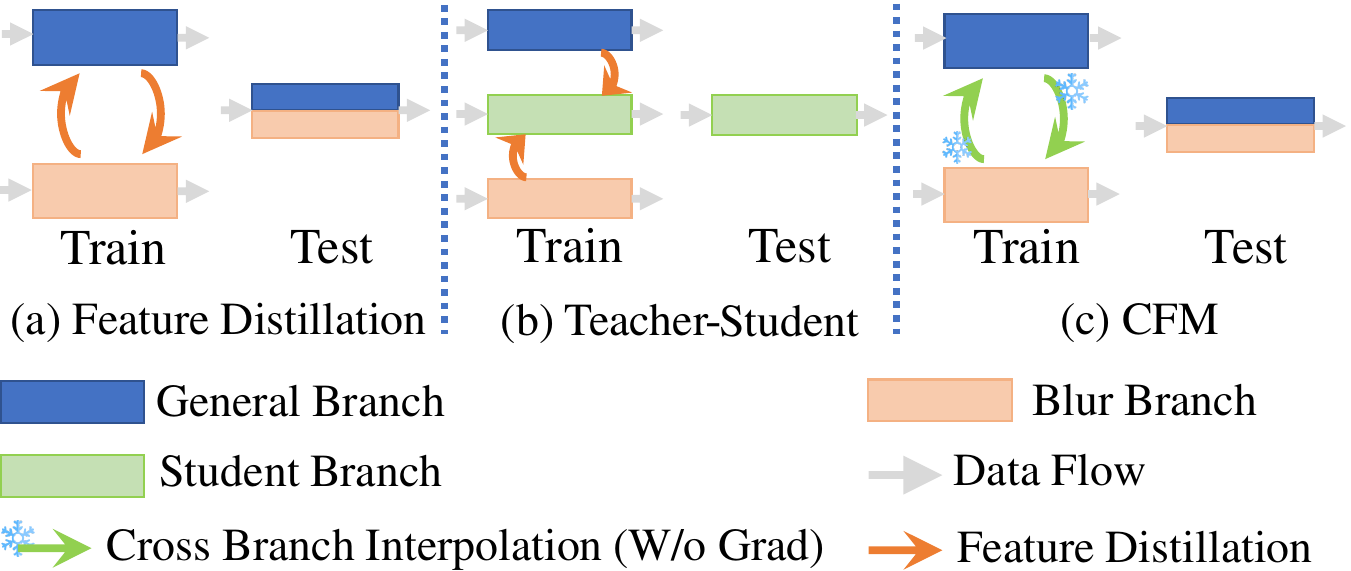}
    % \vspace{-0.6cm}
    \caption{Different fusion strategies.}
    \label{fig:abl_cfm}
    % \vspace{-0.2cm}
\end{minipage}
\begin{minipage}{0.49\linewidth}
    \centering
    % \vspace{0pt}
    \tabcaption{Comparison of different fusion strategies and proportion of branch fusion. 'BP' denotes back propagation.}
        %  \vspace{-0.2cm}
        \label{tab:abl_cdm}
    \resizebox{\linewidth}{!}{
    \setlength{\tabcolsep}{0.5mm}{
        \begin{tabular}{c|c|cc}
            \hline
                \multirow{2}*{Fusion Strategy} & \multirow{2}*{Extra BP} & \multicolumn{2}{c}{LPIPS}  \\ \cline{3-4}
                ~ &~ & Blur & General \\ \hline
                Feature Distillation & + & 0.3674 & 0.3883  \\
                Teacher-Student & + & 0.3760 & 0.3975  \\ 
                % CFM & 0.3837 & 0.3964  \\ 
                % CFM & 0.3664 & 0.3886  \\ 
                CFM & - & \textbf{0.3564} & \textbf{0.3826}  \\ 
                \hline
            \end{tabular}
    }
    }
    \label{tab:abl_cfm}
        
    \end{minipage}

    % \caption{Prediction using different models.}
    % \vspace{-0.6cm}
\end{figure*}

To validate the efficacy of the Cross Fusion Model (CFM), we conducted a comparative analysis of various fusion strategies and the fusion proportion. First, as illustrated in \cref{fig:abl_cfm} and \cref{tab:abl_cfm}, we compared prevalent fusion strategies, encompassing cross feature distillation, the dual-teacher-student framework, and our CFM. 
\cref{tab:abl_cfm} shows that the CFM achieves a 0.006 to 0.008 improvement in LPIPS across all data types compared to the other strategies, without incurring additional computational costs for extra loss and gradient backpropagation, as shown in \cref{fig:abl_cfm}. This shows the suitability of the model interpolation strategy for SR tasks, enabling CFM to maintain optimal fusion performance with minimal computational loss.

\subsection{Discussion about the GAN Artifact Correction (GAC) Methods}
To our knowledge, there is some current work~\cite{jie2022LDL,xie2023desra} aimed at mitigating artifacts in the inference stage of GAN-based image restoration techniques~\cite{realesrgan_wang2021real,liang2021swinir}, aiming to correct the unnatural appearance of generated textures for greater uniformity and consistency. Notwithstanding their notable achievements in texture correction, they do not pay much attention to the understanding of the disparities between blurred and unblurred data distributions. Consequently, they often neglect to assess the necessity for clear and sharp textures in specific regions, especially the blur regions, still suffering from problems such as edge oversharpening and the generation of inappropriate textures in blurred areas. Essentially, GAN artifact correction strategies and blur image super-resolution pursue divergent objectives: the former seeks to refine the synthesis ed texture, while the latter must discern the appropriateness and necessity of texture synthesis across different image regions, especially in the context of blur data. 

\begin{figure}[b]
    \centering
    % \vspace{-0.2cm}
    \includegraphics[width=0.99\linewidth]{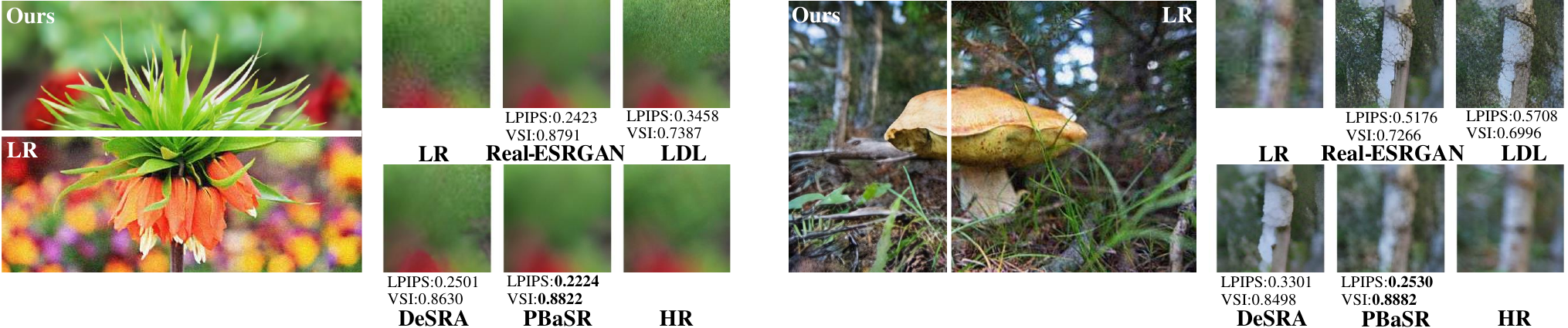}
    % \vspace{-0.5cm}
    \caption{Visual comparison of PBaSR with recent GAN Artifact Correction methods.}    
    % \vspace{-0.5cm}
    \label{fig:ldl_cmp}
\end{figure}

For an intuitive comparison, in \cref{tab:gan_cmp}, we compare our PBaSR framework with prominent GAC methods, including LDL~\cite{jie2022LDL} and DeSRA~\cite{xie2023desra}, using Real-ESRGAN~\cite{realesrgan_wang2021real} as the anchor method. As shown, while GAC methods show slight enhancements on blur data, PBaSR achieves a 0.007$\sim$0.02 LPIPS improvement on blur data over LDL and DeSRA, alongside the best performance on general data. Additionally, we show the real sample comparison in \cref{fig:ldl_cmp}. For instance, as depicted on the left side of \cref{fig:ldl_cmp}, LDL and DeSRA struggle with noise (upper right side) and oversharpening (bottom left side) in blurred regions, while our PBaSR adeptly denoises and preserves defocus. Similarly, the sample on the right side of \cref{fig:ldl_cmp} shows that despite LDL and DeSRA's natural texture synthesis, the over-texturized backgrounds in these examples detract from the blurred region's function of foreground emphasis, compromising overall visual quality. In contrast, PBaSR effectively preserves the blurred region's role in highlighting the subject, thereby enhancing the overall image quality.

In summary, GAN artifact correction methods and blur image blind super-resolution target different objectives—the former improving the synthesized texture and the latter paying more attention to the necessity of texture synthesis in various regions. Integrating the methodologies of these two objectives may be a promising avenue for enhancing outcomes, which we will probably investigate in future research.
\begin{table}[!ht]
    \caption{Comparison of PBaSR with recent GAN Artifact Correction methods on differnt benchmarks. The best is in bold.}
    \centering
    \resizebox{\linewidth}{!}{
    \setlength{\tabcolsep}{3mm}{
        \large
    \begin{tabular}{c|c|cccccc}
    \hline
        Data & Method & LPIPS~$\downarrow$ & AHIQ~$\uparrow$ & DISTS~$\downarrow$ & VIF~$\uparrow$ & GMSD~$\downarrow$ & VSI~$\uparrow$ \\ \hline
        \multirow{4}*{\makecell[c]{Defocus \\ Blur}} & Real-ESRGAN (2021) & 0.4199 & 0.2215 & 0.2313 & 0.0842 & 0.1900 & 0.9547 \\ 
        ~ & LDL$_{\mathrm{ESRGAN}} (2022)$ & 0.4333 & 0.2111 & 0.2434 & 0.0827 & 0.1909 & 0.9538 \\ 
        ~ & DeSRA$_{\mathrm{ESRGAN}} (2023)$ & 0.4186 & 0.2034 & 0.2416 & 0.0871 & 0.1865 & 0.9569 \\ 
        ~ & PBaSR$_{\mathrm{ESRGAN}}$ & \best{0.3986} & \best{0.2386} & \best{0.1952} & \best{0.0907} & \best{0.1773} & \best{0.9609} \\ \hline
        \multirow{4}*{\makecell[c]{Motion \\ Blur}} & Real-ESRGAN (2021) & 0.4104 & 0.2920 & 0.2212 & 0.1135 & 0.1731 & 0.9733 \\ 
        ~ & LDL$_{\mathrm{ESRGAN}}$ (2022)& 0.4161 & 0.2870 & 0.2283 & 0.1133 & 0.1722 & 0.9724 \\ 
        ~ & DeSRA$_{\mathrm{ESRGAN}} (2023)$ & 0.3868  & 0.2686 & 0.2193 & 0.1172 & 0.1665 & 0.9741 \\ 
        ~ & PBaSR$_{\mathrm{ESRGAN}}$ & \best{0.3791} & \best{0.2986} & \best{0.1907} & \best{0.1244} & \best{0.1648} & \best{0.9771} \\ \hline
        \multirow{4}*{\makecell[c]{General }} & Real-ESRGAN (2021) & 0.4739 & 0.2175 & 0.2712 & 0.0506 & 0.2276 & 0.9421 \\ 
        ~ & LDL$_{\mathrm{ESRGAN}}$ (2022)& 0.4784 & 0.2096 & 0.2791 & 0.0502 & 0.2304 & 0.9406 \\ 
        ~ & DeSRA$_{\mathrm{ESRGAN}}$ (2023)& 0.5297 & 0.1854 & 0.3348 & 0.0515 & 0.2346 & 0.9417 \\ 
        ~ & PBaSR$_{\mathrm{ESRGAN}}$ & \best{0.4390} & \best{0.2348} & \best{0.2315} & \best{0.0562} & \best{0.2108} & \best{0.9541} \\ \hline

    \end{tabular}
    }
    }
    \label{tab:gan_cmp}
\end{table}

\subsection{More Comparison Results}
\paragraph{More Quantitative Results}
For reference, we also provided performance comparison between PBaSR and the state-of-the-art methods on more quantitative metrics, including the PSNR, SSIM, CKDN~\cite{ckdn}, STLPIPS~\cite{ghildyal2022stlpips} and TOPIQ-FR~\cite{chen2023topiq}. The results are shown in \cref{tab:pbasr_sota}. Additionally, we also provide the detailed comparison results of different methods on each general BSR benchmarks in \cref{tab:detailed_general_cmp}.

\begin{table}[hbp]
    
    \centering
    % \vspace*{-0.2cm}
    \caption{Extra Quantitative comparison of the proposed method with state-of-the-art methods on Blur data and General data. The \best{best} and \second{second-best} results are bolded in black and red.}
    \resizebox{\linewidth}{!}{
    \setlength{\tabcolsep}{0.5mm}{
        \large
    
    \begin{tabular}{c|c|ccccccc|ccc}
        \hline
        \multirow{2}*{Data} & \multirow{2}*{Metric} & \multirow{2}*{\makecell[c]{SwinIR \\ (2021)}} & \multirow{2}*{\makecell[c]{Real-ESRGAN \\ (2021)}} & \multirow{2}*{\makecell[c]{MM-RealSR \\ (2022)}} & \multirow{2}*{\makecell[c]{FeMaSR  \\ (2022)}}& \multirow{2}*{\makecell[c]{CAL-GAN \\ (2023)}} & \multirow{2}*{\makecell[c]{HAT \\ (2023)}} &\multirow{2}*{\makecell[c]{SRFormer \\ (2023)}} & \multirow{2}*{${\mathrm{\networkname_{ESRGAN}}}$} & \multirow{2}*{${\mathrm{\networkname_{FeMaSR}}}$} & \multirow{2}*{${\mathrm{\networkname_{SRFormer}}}$} \\
        ~ & ~ &~ &~ &~ &~ &~ &~ &~ &~ &~ &~ \\ \hline 

        \multirow{5}*{\makecell[c]{Defocus \\ Blur}}    & PSNR $\uparrow$ & 23.96 & 24.07 & 23.62 & 23.66 & \second{24.41} & 23.61 & 24.24 & \best{24.57} & 24.28 & 24.11 \\
        ~ & SSIM $\uparrow$ &   0.6849 & \second{0.6994} & 0.6847 & 0.6553 & \best{0.7146} & 0.6374 & 0.6920 & 0.6960 & 0.6854 & 0.6686 \\ 
        ~ &  CKDN $\uparrow$ & 0.4870 & 0.4716 & 0.4778 & 0.4426 & 0.4263 & 0.4900 & 0.4686 & 0.4934 & \best{0.5225} & \second{0.5020} \\ 
        ~ & STLPIPS $\downarrow$  & 0.2827 & 0.3016 & 0.2919 & 0.3351 & 0.3044 & 0.2822 & 0.2753 & 0.2820 & \best{0.2442} & \second{0.2507} \\ 
        ~ & TOPIQ-FR $\uparrow$& 0.2467 & 0.2375 & 0.2325 & 0.2072 & 0.2347 & 0.2660 & 0.2533 & 0.2542 & \best{0.2859} & \second{0.2825} \\ \hline
        % ~ & ~ &  ~ & ~ & ~ & ~ & ~ & ~ & ~ & ~ & ~ \\  
    % \hline
        \multirow{5}*{\makecell[c]{Motion \\ Blur}} & PSNR $\uparrow$ & 25.48 & 25.62 & 25.41 & 24.99 & 25.18 & \second{26.23} & 26.11 & \best{26.39} & 26.16 & 26.00 \\ 
        ~ & SSIM $\uparrow$ &   0.7645 & 0.7876 & 0.7845 & 0.6862 & 0.7451 & \best{0.8122} & 0.7752 & \second{0.7934} & 0.7759 & 0.7523 \\
        ~ & CKDN $\uparrow$ & \best{0.5490 }& 0.5379 & 0.5448 & 0.5296 & 0.5110 & 0.5028 & 0.5310 & \second{0.5462} & 0.5445 & 0.5413 \\ 
        ~ & STLPIPS $\downarrow$ & 0.2300 & 0.2174 & 0.2157 & 0.3007 & 0.2777 & 0.2111 & 0.2179 & \best{0.1987} & \second{0.2013} & 0.2258 \\
        ~ & TOPIQ-FR $\uparrow$ & 0.2608 & 0.2581 & 0.2419 & 0.2585 & 0.2190 & 0.2494 & 0.2638 & 0.2660 & \best{0.2896} & \second{0.2802} \\ \hline
        % ~ & ~ &  ~ & ~ & ~ & ~ & ~ & ~ & ~ & ~ & ~ \\ 
    % \hline
        \multirow{5}*{DIV2K}                          & PSNR $\uparrow$ & 21.25 & 20.79 & 21.37 & 20.16 & \best{21.91} & 21.2 & \second{21.88} & 21.82 & 21.74 & 21.64 \\
        ~ & SSIM $\uparrow$ &  0.5604 & 0.5425 & 0.5648 & 0.5045 & \best{0.5789} & 0.5321 & \second{0.5788} & 0.5655 & 0.5648 & 0.5639 \\ 
        ~ & CKDN $\uparrow$ & 0.5123 & 0.4840 & 0.4784 & 0.4691 & 0.4498 & 0.4931 & 0.4964 & \second{0.5144} & \best{0.5218} & 0.5103 \\
        ~ & STLPIPS $\downarrow$ & 0.3224 & 0.3629 & 0.4042 & 0.3727 & 0.4222 & \best{0.2837} & 0.3436 & 0.3476 & \second{0.2874} & 0.3018 \\ 
        ~ & TOPIQ-FR $\uparrow$ & 0.2437 & 0.2216 & 0.2107 & 0.1997 & 0.2161 & 0.2672 & 0.2451 & 0.2501 & \best{0.2834} & \second{0.2760} \\ 
        % ~ & ~ &  ~ & ~ & ~ & ~ & ~ & ~ & ~ & ~ & ~ \\ 
    \hline
    \end{tabular}
    }}
    
    \label{tab:pbasr_sota}
\end{table}

\paragraph{More Qualitative Results}
We provide more visual comparison results between PBaSR and the state-of-the-art methods in \cref{fig:qualitative} and \cref{fig:qualitative_2}.

\begin{figure*}[t]
    \centering
    \includegraphics[width=\linewidth]{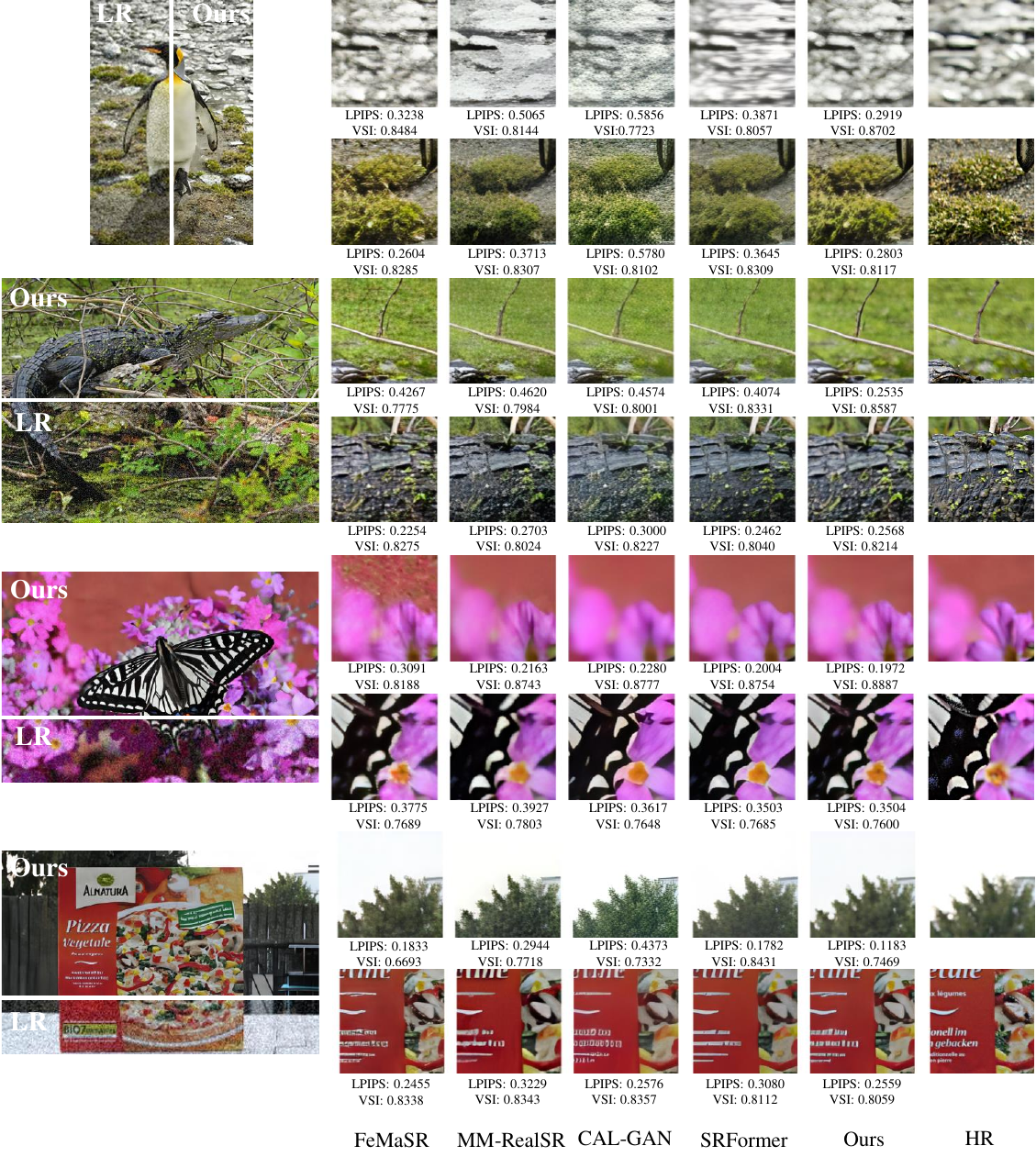}
    \caption{More visual comparison of the proposed method with state-of-the-art methods on blur data.}
    \label{fig:qualitative}
\end{figure*}
\begin{figure*}
    \centering
    \includegraphics[width=\linewidth]{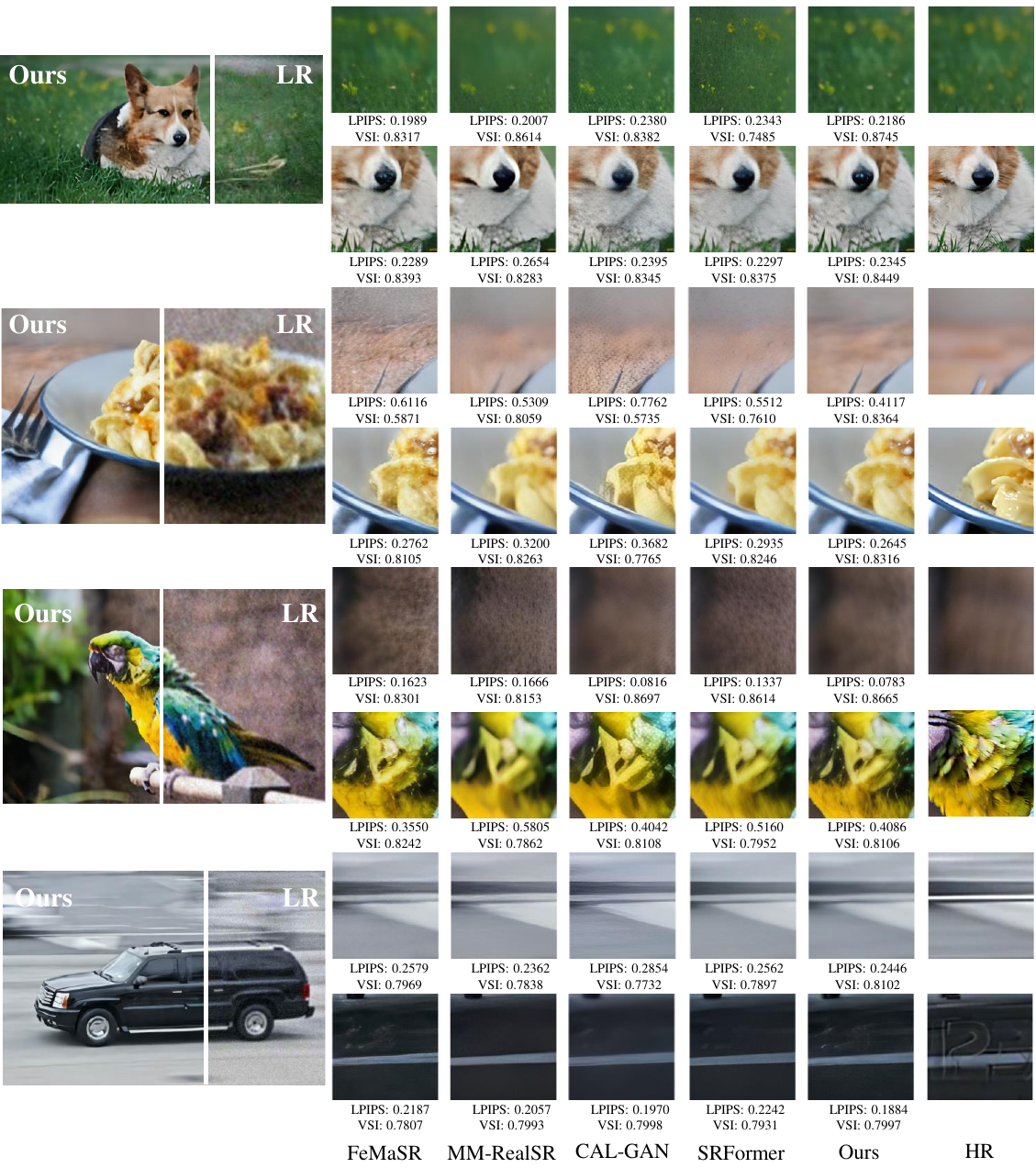}
    \caption{More Visual comparison of the proposed method with state-of-the-art methods on blur data.}
    \label{fig:qualitative_2}
\end{figure*}

\begin{table}[!ht]
    \caption{Detailed metrics comparison with state-of-the-art methods on each general BSR benchmarks. The \best{best} and \second{second-best} results are bolded in black and red.}
    \centering
    \resizebox{\linewidth}{!}{
        \setlength{\tabcolsep}{0.5mm}{
            \large
    \begin{tabular}{c|c|ccccccc|ccc}
    \hline
    \multirow{2}*{Data} & \multirow{2}*{Metric} & \multirow{2}*{\makecell[c]{SwinIR \\ (2021)}} & \multirow{2}*{\makecell[c]{Real-ESRGAN \\ (2021)}} & \multirow{2}*{\makecell[c]{MM-RealSR \\ (2022)}} & \multirow{2}*{\makecell[c]{FeMaSR  \\ (2022)}}& \multirow{2}*{\makecell[c]{CAL-GAN \\ (2023)}} & \multirow{2}*{\makecell[c]{HAT \\ (2023)}} &\multirow{2}*{\makecell[c]{SRFormer \\ (2023)}} & \multirow{2}*{${\mathrm{\networkname_{ESRGAN}}}$} & \multirow{2}*{${\mathrm{\networkname_{FeMaSR}}}$} & \multirow{2}*{${\mathrm{\networkname_{SRFormer}}}$} \\
    ~ & ~ &~ &~ &~ &~ &~ &~ &~ &~ &~ &~ \\ \hline 
    \multirow{6}*{\makecell[c]{Urban100}} & LPIPS~$\downarrow$& 0.4248 & 0.4994 & 0.5373 & 0.4234 & 0.5137 & 0.5078 & 0.4249 & 0.4675 & \second{0.4155} & \best{0.3982} \\  
        ~ & AHIQ~$\uparrow$ & 0.1913 & 0.1675 & 0.1640 & 0.1638 & 0.1413 & 0.1847 & \second{0.2094} & 0.1734 & 0.1918 & \best{0.2245} \\  
        ~ & DISTS~$\downarrow$ & \second{0.2501} & 0.3056 & 0.3394 & 0.2709 & 0.3174 & 0.3235 & 0.2587 & 0.2955 & 0.2746 & \best{0.2390} \\  
        ~ & VIF~$\uparrow$ & 0.0724 & 0.0622 & 0.0608 & 0.0662 & 0.0594 & 0.0676 & \best{0.0776} & 0.0669 & 0.0728 & \second{0.0766} \\  
        ~ & GMSD~$\downarrow$ & 0.2659 & 0.2832 & 0.2855 & \second{0.2494} & 0.2710 & 0.2820 & 0.2653 & 0.2671 & 0.2558 & \best{0.2481} \\  
        ~ & VSI~$\uparrow$ & 0.9028 & 0.8871 & 0.8849 & \second{0.9146} & 0.8966 & 0.8864 & 0.9049 & 0.9037 & 0.9131 & \best{0.9159} \\ \hline
    \multirow{6}*{\makecell[c]{BSDS100}}  & LPIPS~$\downarrow$ & 0.4691 & 0.5176 & 0.5410 & 0.4479 & 0.5210 & 0.5411 & 0.4987 & 0.5022 & \second{0.4387} & \best{0.4442}\\  
        ~ & AHIQ~$\uparrow$ & 0.1557 & 0.1711 & 0.1536 & 0.1862 & 0.1386 & 0.1760 & 0.1641 & 0.1673 & \best{0.2016} & \second{0.1998} \\  
        ~ & DISTS~$\downarrow$ & 0.3067 & 0.3387 & 0.3382 & \best{0.2612} & 0.3513 & 0.3625 & 0.3240 & 0.2910 & \second{0.2674} & 0.2687 \\  
        ~ & VIF~$\uparrow$ & 0.0428 & 0.0394 & 0.0394 & 0.0438 & 0.0400 & 0.0407 & 0.0437 & 0.0448 & \best{0.0460} & \second{0.0454} \\  
        ~ & GMSD~$\downarrow$ & 0.2205 & 0.2307 & 0.2333 & \best{0.2075} & 0.2185 & 0.2375 & 0.2264 & 0.2146 & 0.2104 & \second{0.2079} \\  
        ~ & VSI~$\uparrow$ & 0.9001 & 0.8921 & 0.8902 & \best{0.9048} & 0.8977 & 0.8858 & 0.8962 & 0.8989 & 0.9043 & \second{0.9047} \\ \hline
    \multirow{6}*{\makecell[c]{manga109}}  & LPIPS~$\downarrow$ & 0.3714 & 0.3968 & 0.4194 & \best{0.3443} & 0.4007 & 0.4175 & 0.3744 & 0.3849 & \second{0.3346} & 0.3488 \\  
        ~ & AHIQ~$\uparrow$ & 0.2032 & 0.2010 & 0.1649 & 0.1880 & 0.1542 & 0.1704 & 0.1972 & 0.1743 & \best{0.2230} & \second{0.2159} \\  
        ~ & DISTS~$\downarrow$ & 0.2001 & 0.2225 & 0.2448 & 0.1956 & 0.2200 & 0.2458 & 0.2117 & 0.2185 & \best{0.1851} & \second{0.1866} \\  
        ~ & VIF~$\uparrow$ & 0.0850 & 0.0732 & 0.0725 & 0.0799 & 0.0724 & 0.0770 & \best{0.0881} & 0.0776 & \second{0.0867} & 0.0862 \\  
        ~ & GMSD~$\downarrow$ & 0.2526 & 0.2641 & 0.2655 & \best{0.2351} & 0.2563 & 0.2590 & 0.2523 & 0.2474 & \second{0.2376} & 0.2391 \\  
        ~ & VSI~$\uparrow$ & 0.9204 & 0.9123 & 0.9099 & \second{0.9284} & 0.9149 & 0.9139 & 0.9218 & 0.9219 & \best{0.9289} & 0.9282 \\ \hline
    \multirow{6}*{\makecell[c]{Set14}}  & LPIPS~$\downarrow$ & 0.4398 & 0.4697 & 0.4755 & 0.4033 & 0.4520 & 0.4667 & 0.4394 & 0.4495 & \best{0.3944} & \second{0.3971} \\  
        ~ & AHIQ~$\uparrow$ & 0.1521 & 0.1600 & 0.1407 & 0.1736 & 0.1393 & 0.1572 & 0.1622 & 0.1486 & \best{0.2017} & \second{0.1841} \\  
        ~ & DISTS~$\downarrow$ & 0.2723 & 0.2814 & 0.2982 & 0.2461 & 0.2803 & 0.3063 & 0.2901 & 0.2639 & \second{0.2439} & \best{0.2343} \\  
        ~ & VIF~$\uparrow$ & 0.0625 & 0.0600 & 0.0605 & 0.0621 & 0.0600 & 0.0622 & \best{0.0663} & 0.0625 & 0.0659 & \second{0.0663} \\  
        ~ & GMSD~$\downarrow$ & 0.2360 & 0.2466 & 0.2464 & \second{0.2198} & 0.2344 & 0.2397 & 0.2336 & 0.2298 & 0.2223 & \best{0.2189} \\  
        ~ & VSI~$\uparrow$ & 0.9028 & 0.8923 & 0.8918 & \best{0.9064} & 0.9021 & 0.8921 & 0.9019 & 0.9034 & \second{0.9062} & 0.9108 \\ \hline
    \multirow{6}*{\makecell[c]{Set5}}  & LPIPS~$\downarrow$ & 0.3998 & 0.4723 & 0.4874 & 0.3611 & 0.4794 & 0.4726 & 0.4679 & 0.3772 & \best{0.3510} & \second{0.3512} \\  
        ~ & AHIQ~$\uparrow$ & \second{0.2125} & \best{0.2227} & 0.1184 & 0.1592 & 0.2005 & 0.1984 & 0.1447 & 0.1850 & 0.2083 & 0.1541 \\  
        ~ & DISTS~$\downarrow$ & 0.3007 & 0.3571 & 0.3668 & \best{0.2636} & 0.3723 & 0.3574 & 0.3562 & 0.2822 & 0.2778 & \second{0.2728} \\  
        ~ & VIF~$\uparrow$ & 0.0628 & 0.0568 & 0.0565 & 0.0586 & 0.0559 & 0.0603 & \second{0.0647} & 0.0614 & 0.0638 & \best{0.0661} \\  
        ~ & GMSD~$\downarrow$ & 0.2255 & 0.2478 & 0.2482 & \best{0.2170} & 0.2365 & 0.2454 & 0.2462 & 0.2258 & \second{0.2203} & 0.2207 \\  
        ~ & VSI~$\uparrow$ & 0.9169 & 0.9094 & 0.9118 & 0.9104 & 0.9057 & 0.9098 & 0.9165 & \best{0.9191} & \second{0.9174} & 0.9171 \\ \hline
    \multirow{6}*{\makecell[c]{DIV2K}}  & LPIPS~$\downarrow$ & 0.4181 & 0.4739 & 0.4929 & \second{0.3848} & 0.4847 & 0.4828 & 0.4170 & 0.4390 & \best{0.3826} & 0.3892 \\  
        ~ & AHIQ~$\uparrow$ & 0.2311 & 0.2175 & 0.2075 & 0.2396 & 0.1888 & 0.2039 & 0.2329 & 0.2348 & \best{0.2621} & \second{0.2579} \\  
        ~ & DISTS~$\downarrow$ & 0.2314 & 0.2712 & 0.3050 & \best{0.1928} & 0.2869 & 0.3186 & 0.2478 & 0.2315 & 0.2059 & \second{0.1966} \\  
        ~ & VIF~$\uparrow$ & 0.0567 & 0.0506 & 0.0504 & 0.0561 & 0.0504 & 0.0519 & 0.0587 & 0.0562 & \best{0.0602} & \second{0.0595} \\  
        ~ & GMSD~$\downarrow$ & 0.2155 & 0.2276 & 0.2303 & \best{0.1976} & 0.2215 & 0.2320 & 0.2164 & 0.2108 & \second{0.2003} & 0.2017 \\  
        ~ & VSI~$\uparrow$ & 0.9487 & 0.9421 & 0.9399 & \second{0.9581} & 0.9484 & 0.9389 & 0.9485 & 0.9541 & \best{0.9590} & 0.9574 \\ \hline
    \multirow{6}*{\makecell[c]{All}}  & LPIPS~$\downarrow$ & 0.4202 & 0.4703 & 0.4952 & 0.3986 & 0.4775 & 0.4850 & 0.4284 & 0.4463 & \best{0.3912} & \second{0.3937} \\  
        ~ & AHIQ~$\uparrow$ & 0.1944 & 0.1890 & 0.1707 & 0.1932 & 0.1557 & 0.1828 & 0.1990 & 0.1860 & \second{0.2190} & \best{0.2223} \\  
        ~ & DISTS~$\downarrow$ & 0.2475 & 0.2839 & 0.3060 & \second{0.2303} & 0.2928 & 0.3115 & 0.2615 & 0.2587 & 0.2331 & \best{0.2229} \\  
        ~ & VIF~$\uparrow$ & 0.0646 & 0.0568 & 0.0563 & 0.0619 & 0.0560 & 0.0598 & \best{0.0674} & 0.0617 & 0.0668 & \second{0.0673} \\  
        ~ & GMSD~$\downarrow$ & 0.2387 & 0.2515 & 0.2536 & \best{0.2225} & 0.2418 & 0.2523 & 0.2402 & 0.2350 & 0.2261 & \second{0.2243} \\  
        ~ & VSI~$\uparrow$ & 0.9176 & 0.9080 & 0.9059 & 0.9257 & 0.9139 & 0.9060 & 0.9174 & 0.9192 & \second{0.9257} & \best{0.9260} \\ \hline
    \end{tabular}
        }
    }
    \label{tab:detailed_general_cmp}
\end{table}
\end{document}